\documentclass[twoside,11pt]{article}

\usepackage{jmlr2e}

\usepackage{algorithm}
\usepackage{algorithmic}
\usepackage{amsmath}

\renewcommand{\algorithmicrequire}{\textbf{Arguments:}}
\long\def\symbolfootnote[#1]#2{\begingroup%
\def\thefootnote{\fnsymbol{footnote}}\footnote[#1]{#2}\endgroup} 

\ShortHeadings{Truncating the Loop Series Expansion for BP}{Vicen\c{c} G\'omez and
  J.M. Mooij and H.J. Kappen}
\firstpageno{1}

\begin{document}

\title{Truncating the Loop Series Expansion for Belief Propagation}

\author{\name Vicen\c{c} G\'omez\thanks{Visiting Radboud University Nijmegen}
\email vgomez@iua.upf.edu \\
\addr Departament de Tecnologies de la Informaci\'o i les Comunicacions\\
Universitat Pompeu Fabra\\
Passeig de Circumval$\cdot$laci\'o 8, 08003 Barcelona, Spain\\\\\
Barcelona Media Centre d'Innovaci\'o\\
Ocata 1, 08003 Barcelona, Spain\\
\AND
\name Joris M. Mooij
\email j.mooij@science.ru.nl \\
\name Hilbert J. Kappen
\email b.kappen@science.ru.nl \\
\addr Department of Biophysics\\
Radboud University Nijmegen\\
6525 EZ Nijmegen, The Netherlands
}

\editor{Michael Jordan}

\maketitle

\begin{abstract}
Recently, \citet{chertkov2006a} derived an exact expression for the partition sum
(normalization constant) corresponding to a graphical model, which is an
expansion around the belief propagation (BP) solution. By adding correction terms to
the BP free energy, one for each ``generalized loop'' in the factor graph, the
exact partition sum is obtained.
However, the usually enormous number of
generalized loops generally prohibits summation over \emph{all} correction
terms.
In this article we introduce truncated loop series BP (TLSBP), a particular way
of truncating the loop series of Chertkov \& Chernyak by considering generalized loops
as compositions of simple loops.
We analyze the performance of TLSBP in different scenarios, including the Ising model
on square grids and regular random graphs, and on Promedas, a large probabilistic
medical diagnostic system.
We show that TLSBP often improves upon the accuracy of the BP solution, at the
expense of increased computation time.
We also show that the performance of TLSBP strongly depends on
the degree of interaction between the variables.
For weak interactions, truncating the series leads to significant improvements,
whereas for strong interactions it can be ineffective,
even if a high number of terms is considered.
\end{abstract}


\section{Introduction}
\label{sec:intro}
Belief propagation \citep{pearl88, murphy99} is a popular inference
method that yields exact marginal probabilities on graphs without
loops and can yield surprisingly accurate results on graphs with
loops.
BP has been shown to outperform other methods in rather diverse and
competitive application areas, such as error correcting codes
\citep{gallager63, mcelice98}, low level vision \citep{freeman00},
combinatorial optimization \citep{mezard02} and stereo vision
\citep{sun05}.

Associated to a probabilistic model is the partition sum, or normalization
constant, from which marginal probabilities can be obtained.
Exact calculation of the partition function is only feasible for small problems,
and there is considerable statistical physics literature devoted to the approximation of this
quantity.
Existing methods include stochastic Monte Carlo techniques \citep{Potamianos1997}
or deterministic algorithms which provide lower bounds \citep{Jordan99, Leisink2001},
upper bounds \citep{wainwright05}, or approximations \citep{yedidia05}.

Recently, \citet{chertkov2006a} have presented a loop series expansion
formula
that computes correction terms to the belief propagation
approximation of the partition sum.
The series consists of a sum over all so-called generalized
loops in the graph. When all loops are taken into account,
Chertkov \& Chernyak show that the exact result is recovered.
Since the number of generalized loops in a graphical model easily
exceeds the number of configurations of the model, one could argue
that the method is of little practical value. However, if one could
truncate the expansion in some principled way, the method could provide
an efficient improvement to BP.\footnote{Note that the number of generalized loops in a
finite graph is finite,
and strictly speaking, the term \emph{series} denotes an \emph{infinite}
sequence of terms.
For clarity, we prefer to use the original terminology.}

Most inference algorithms on loopy graphs can be viewed as generalizations of
BP, where messages are propagated between regions of variables.
For instance, the junction-tree algorithm \citep{lauritzen88} which transforms
the original graph in a region tree such that the influence of all loops in the
original graph is implicitly captured, and the exact result is obtained.
However, the complexity of this algorithm is exponential in time and space
on the size of the largest clique of the resulting join tree, or equivalently,
on the tree-width of the original graph, a parameter which measures the network
complexity.
Therefore, for graphs with high tree-width one is resorted to approximate
methods such as Monte Carlo sampling or generalized belief propagation (GBP)~\citep{yedidia05},
which captures the influence of short loops using regions which contain
them.
One way to select valid regions is the cluster variation method
(CVM)~\citep{Pelizola05}.
In general, selecting a good set of regions is not an easy task, as described
in~\citep{Welling05}.
Alternatively, double-loop methods can be used
\citep{hes03a,yuille02} which are guaranteed to converge, often at the
cost of more computation time.

In this work we propose TLSBP, an algorithm to compute generalized
loops in a graph which are then used for the approximate computation of
the partition sum and the single-node marginals.
The proposed algorithm is parametrized by two arguments which are used to prune
the search for generalized loops.
For large enough values of these parameters, all generalized loops
present in a graph are retrieved and the exact result is obtained.
One can then study how the error is progressively corrected as more
terms are considered in the series.
For cases were exhaustive computation of all loops is not feasible,
the search can be pruned, and the result is a truncated approximation
of the exact solution.
We focus mainly on problems where BP converges easily, without the
need of damping or double loop alternatives \citep{hes03a,yuille02}
to force convergence.
It is known that accuracy of the BP solution and convergence rate
are negatively correlated. Throughout the paper we show evidence that for those
cases where BP has difficulties to converge, loop corrections are of little
use, since loops of all lengths tend to have contributions of similar order of magnitude.

The paper is organized as follows. In Section~\ref{sec:method}
we briefly summarize the series expansion method of
\citet{chertkov2006a}. In Section~\ref{sec:charac} we provide a
formal characterization of the different types of generalized loops
that can be present in an arbitrary graph.
This description is relevant to understand the proposed algorithm
described in Section~\ref{sec:algor}.
We present experimental results in Section~\ref{sec:experim} for the
Ising model on grids, regular random graphs and medical diagnosis.
Concering grids and regular graphs, we show that the success of restricting
the loop series expansion to a reduced quantity of loops depends on the type
of interactions between the variables in the network.
For weak interactions, the largest correction terms come from the small
elementary loops and therefore truncation of the series at some
maximal loop length can be effective. For strong interactions, loops of
all lengths contribute significantly and truncation is of limited
use. We numerically show that when more loops are taken into account,
the error of the partition sum decreases and when all loops are taken
into account the method is correct up to machine precision.
We also apply the truncated loop expansion to a large probabilistic medical
diagnostic decision support system \citep{wie99b}.
The network has 2000 diagnoses and about 1000 findings
and is intractable for computation. However, for each patient case
unobserved findings and irrelevant diagnoses can be pruned from the
network. This leaves a much smaller network that may or may not be
tractable depending on the set of clamped findings. For a number of
patient cases, we compare the BP approximation and the truncated loop
correction. We show results and characterize when the loop corrections
significantly improve the accuracy of the BP solution.
Finally, in Section~\ref{sec:discussion} we provide some concluding
remarks.

\section{BP and the Loop Series Expansion}
\label{sec:method}
Consider a probability model on a set of binary variables $x_i=\pm 1,
i=1,\ldots,n$:
\begin{align}\label{eq:prob}
P(x)=\frac{1}{Z}\prod_{\alpha=1}^m f_{\alpha}(x_\alpha),\qquad Z=\sum_x
\prod_{\alpha=1}^m f_\alpha(x_\alpha),
\end{align}
where $\alpha=1,\ldots,m$ labels interactions (factors) on subsets of
variables $x_\alpha$, and $Z$ is the partition function, which sums
over all possible states or variable configurations.
Note that the only restriction here is that variables are binary, since
arbitrary factor nodes are allowed, as in \citep{chertkov2006a}.

The probability distribution in \eqref{eq:prob} can be directly expressed by
means of a factor graph \citep{kschischang01factor}, a bipartite graph
where variable nodes $i$ are connected to factor nodes $\alpha$
if and only if $x_i$ is an argument of $f_\alpha$.
Figure \ref{fig:ising4x4} (left) on page \pageref{page-ref} shows an
example of a graph where variable and factor nodes are indicated
by circles and squares respectively.


For completeness, we now briefly summarize Pearl's belief propagation (BP)
\citep{pearl88} and define the Bethe free energy.
If the graph is acyclic, BP iterates the following message update
equations, until a fixed point is reached:
\begin{align*}
\textbf{variable $i$ to factor $\alpha$:}& & \mu_{i\rightarrow
  \alpha}(x_i) & = \prod_{\beta \ni i \setminus \{\alpha\}}{
  \mu_{\beta \rightarrow i} (x_i) } ,\\
\textbf{factor $\alpha$ to variable $i$:}& & \mu_{\alpha \rightarrow
  i}(x_i) & = \sum_{x_{\alpha \setminus \{i\}}}{f_{\alpha}(x_{\alpha})
  \prod_{j \in \alpha \setminus \{i\}}{ \mu_{j \rightarrow \alpha}
    (x_j) }} ,
\end{align*}
where $i \in \alpha$ denotes variables included in factor $\alpha$, and
$\alpha \ni i$ denotes factor indices $\alpha$ which have $i$ as argument.
After the fixed point is reached, exact marginals and correlations associated
with the factors (``beliefs'') can be computed using:
\begin{align*}
b_i(x_i) & \propto \prod_{\alpha \ni i}{\mu_{\alpha \rightarrow i} (x_i) } ,\\
b_\alpha(x_\alpha) & \propto f_\alpha(x_\alpha) \prod_{i \in \alpha}
{\mu_{i \rightarrow \alpha} (x_i) } ,
\end{align*}
where $\propto$ indicates normalization so that beliefs sum to one.

For graphs with cycles the same update equations can be iterated
(the algorithm is then called loopy, or iterative, belief propagation),
and one can still obtain very accurate approximations of the beliefs.
However, convergence is not guaranteed in these cases.
For example, BP can get stuck in limit cycles.
An important step towards the understanding and characterization
of the convergence properties of BP came from the observation that fixed
points of this algorithm correspond to stationary points of a particular function of 
the beliefs, known as the Bethe free energy~\citep{yedidia00}, which is
defined as:
\begin{align}\label{eq:Bethe}
F_{BP} & = U_{BP} - H_{BP},
\end{align}
where $U_{BP}$ is the Bethe average energy:
\begin{align*}
U_{BP} &= -\sum_{\alpha=1}^{m}{\sum_{x_\alpha}{b_\alpha(x_\alpha) \log f_\alpha(x_\alpha)} },
\end{align*}
and $H_{BP}$ is the Bethe approximate entropy:
\begin{align}\label{eq:entropy}
H_{BP} &= -\sum_{\alpha=1}^{m}{\sum_{x_\alpha}{b_\alpha(x_\alpha) \log b_\alpha(x_\alpha)} }
+ \sum_{i=1}^{n}{(d_i-1) \sum_{x_i}{b_i(x_i)\log b_i(x_i)} }.
\end{align}
where $d_i$ is the number of neighboring factor nodes of variable node $i$.
The second term in \eqref{eq:entropy} ensures that every node in the graph is counted
once, see~\citep{yedidia05} for details.
The BP algorithm tries to minimize \eqref{eq:Bethe} and, for trees,
the exact partition function can be obtained after the fixed point has been
reached, $Z = \exp(-F_{BP})$.
However, for graphs with loops $F_{BP}$ provides just an approximation.


If one can calculate the exact partition function $Z$ defined in Equation~\eqref{eq:prob},
one can also calculate any marginal in the network. For instance, the marginal
\begin{align*}
P_i({x_i}) & = \left.\frac{\partial \log Z(\theta_i)}{\partial
    \theta_i(x_i)}\right|_{\theta_i\rightarrow 0} \qquad \text{where} \qquad Z(\theta_i) := \sum_x e^{\theta_i x_i} \prod_{\alpha=1}^m f_\alpha(x_\alpha)
\end{align*}
is the partition sum of the network, perturbed by an additional
local field potential $\theta_i$ on variable $x_i$.

Alternatively, one can compute different partition functions for different
settings of the variables, and derive the marginals from ratios of them:
\begin{align}\label{eq:Zclamp}
P_i(x_i) & = \frac{Z^{x_i}}{\displaystyle \sum_{x'_i}{Z^{x'_i}} }
\end{align} 
where $Z^{x_i}$ indicates the partition function calculated from the same model
conditioning on variable $i$, i.e, with variable $i$ fixed (clamped) to value $x_i$.
Therefore, approximation errors in the computation of any marginal can
be related to approximation errors in the computation of $Z$. We will
thus focus on the approximation of $Z$ mainly, although marginal
probabilities will be computed as well.

Of central interest in this work is the concept of generalized loop,
which is defined in the following way:

\begin{definition}\label{def:gen loop}
A \textbf{generalized loop} in a graph $G=\langle V, E\rangle$ is any
subgraph $C = \langle V', E'\rangle$, $V' \subseteq V, E' \subseteq (V'\times V') \cap E$
such that each node in $V'$ has degree two or larger.
The length (size) of a generalized loop is its number of edges.
\end{definition}

For the rest of the paper, the terms loop and generalized loop are used
interchangeably.
The main result of \citep{chertkov2006a} is the following.
Let $b_\alpha(x_\alpha), b_i(x_i)$ denote the beliefs after the BP algorithm has
been converged, and let $Z_\mathrm{BP}=\exp(-F_\mathrm{BP})$ denote the
corresponding approximation to the partition sum, with $F_\mathrm{BP}$
the value of the Bethe free energy evaluated at the BP solution.
Then $Z_\mathrm{BP}$ is related to the exact partition sum $Z$ as:
\begin{align}\label{eq:series}
Z&=Z_\mathrm{BP}\left(1+\sum_{C\in\mathcal{C}} r(C)\right) , &
r(C)&=\prod_{i\in C} \mu_i(C) \prod_{\alpha\in C}\mu_\alpha(C)
\end{align}
where summation is over the set $\mathcal{C}$ of all generalized loops
in the factor graph.
Any term $r(C)$ in the series corresponds to a product with as many
factors as nodes present in the loop.
Each factor is related to the beliefs at each variable node
or factor node according to the following formulas:
\begin{align}\label{eq:term-var}
\mu_i(C)&=\frac{(1-m_i)^{q_i(C)-1}+(-1)^{q_i(C)}(1+m_i)^{q_i(C)-1}}{2(1-m_i^2)^{q_i(C)-1}},
& q_i(C)&=\sum_{\alpha \in C,\alpha\ni i} 1\\\label{eq:term-fact}
\mu_\alpha(C) & = \sum_{x_\alpha}{b_{\alpha}(x_\alpha) \prod_{i \in C, i \in
    \alpha}{(x_i - m_i)}},
\end{align}
where $m_i=\sum_{x_i}b_i(x_i)x_i = b_i(+) - b_i(-)$ is the expected value of $x_i$
computed in the BP approximation.
Generally, terms $r(C)$ can take positive or negative values.
Even the same variable $i$ may have positive or negative subterms $\mu_i$
depending on the structure of the particular loop.

Expression \eqref{eq:series} represents an exact and finite
decomposition of the partition function with the first term of the
series being exactly represented by the BP solution.
Note that, although the series is finite, the number of generalized
loops in the factor graph can be enormous
and easily exceed 
the number of configurations $2^n$.
In these cases the loop series is less efficient than the
most naive way to compute $Z$ exactly, namely by summing the
contributions of all $2^n$ configurations one by one. 

On the other hand, it may be that restricting the sum in \eqref{eq:series}
to a subset of the total generalized loops captures the most important
corrections and may yield a significant improvement in comparison to
the BP estimate.
We therefore define the truncated form of the loop corrected partition
function as:
\begin{align}\label{eq:Zlc}
Z_{TLSBP}&=Z_\mathrm{BP}\left(1+\sum_{C \in \mathcal{C'}} r(C)\right)
\end{align}
where summation is over the subset $\mathcal{C'} \subseteq \mathcal{C}$
obtained by Algorithm \ref{al:TLSBP}, which we will discuss in Section
\ref{sec:algor}.
Approximations for the single-node marginals can then be obtained
from~\eqref{eq:Zlc}, using the method proposed in Equation~\eqref{eq:Zclamp}:
\begin{align}\label{eq:Zclamp-tlsbp}
b'_i(x_i) & = \frac{Z^{x_i}_{TLSBP}}{\displaystyle \sum_{x'_i}{Z^{x'_i}_{TLSBP}} }
\end{align}
Because the the terms $r(C)$ can have different signs, the approximation $Z_{TLSBP}$ 
is in general not a bound of the exact $Z$, but just an approximation.



\section{Loop Characterization}
\label{sec:charac}
In this section we 
characterize different types of generalized loops that can be present
in a graph.
This classification is the basis of the algorithm described in the
next section and also exemplifies the different shapes a generalized
loop can take.
For clarity, we illustrate them by means of a factor graph arranged in
a square lattice with only pairwise interactions. However,
definitions are not restricted to this particular model and can be
applied generally to any factor graph.
\begin{figure}
\begin{center}
\includegraphics[scale=.4]{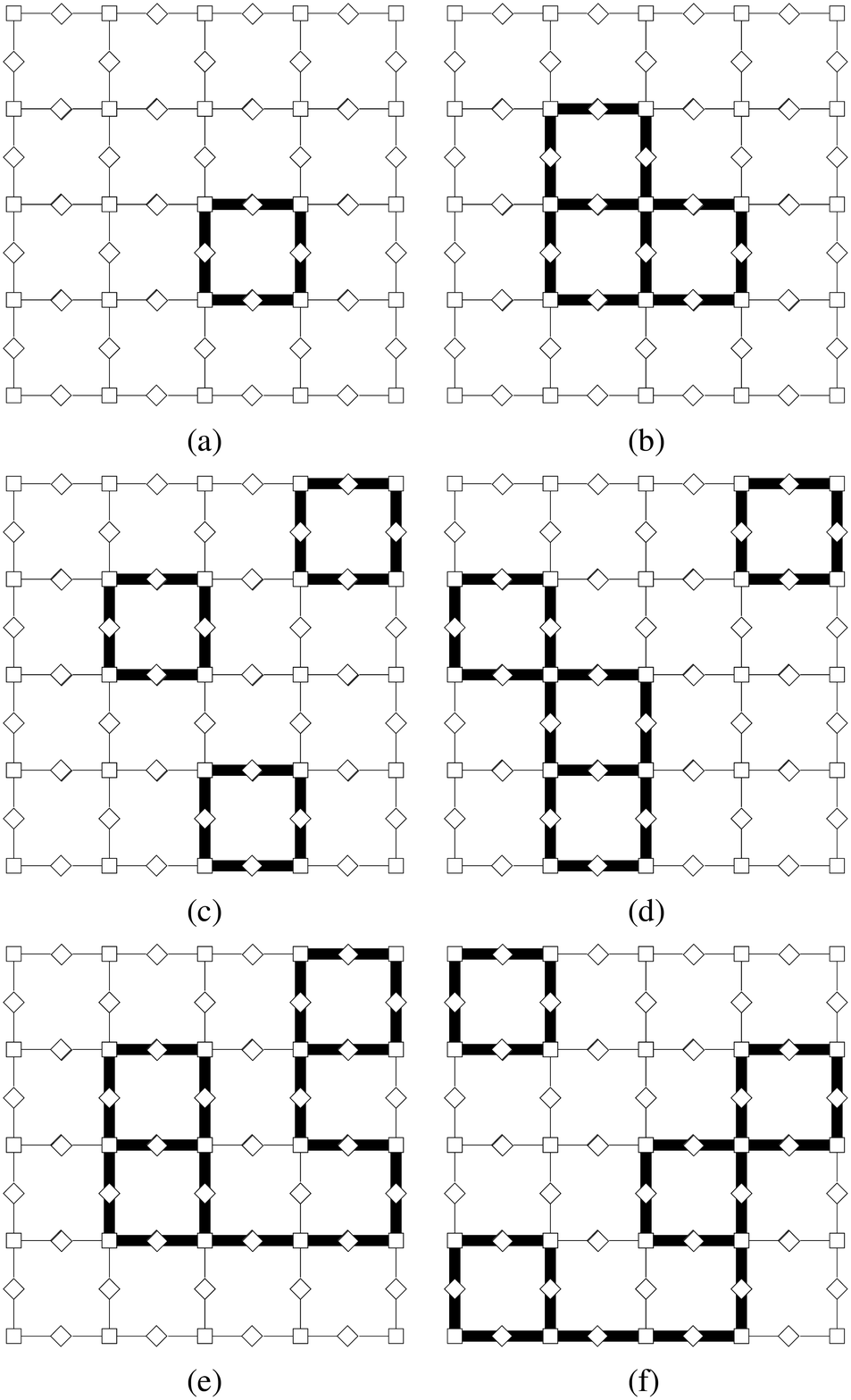}
\end{center}
\caption{
Examples of generalized loops in a factor graph with lattice
structure.
Variable nodes and factor nodes are represented
as squares and rhombus respectively. Generalized loops are indicated
using bold edges underlying the factor graph. \textbf{(a)} A
simple loop. \textbf{(b)} A non-simple loop which is neither a disconnected loop
nor a complex loop. \textbf{(c)} A disconnected loop of three
components, each a simple loop. \textbf{(d)} A disconnected loop of two
components, the left one a non-simple loop. \textbf{(e)} A complex loop which
is not a disconnected loop. \textbf{(f)} A complex loop which is also a
disconnected loop. \emph{(See text for definitions)}. }
\label{fig:ising loop}
\end{figure}
\begin{definition}\label{def:sloop}
A simple (elementary) generalized loop (from now on
\textbf{simple loop}) is defined as a connected subgraph of the
original graph where all nodes have exactly degree
two. \end{definition}

This type of generalized loop coincides with the concept of simple
circuit or simple cycle in graph theory: a path which starts and ends
at the same node with no repeated vertices except for the start and
end vertex. Figure~\ref{fig:ising loop}a shows an example of a
simple loop of size 8.
On the contrary, in Figure~\ref{fig:ising loop}b we show an example of
generalized loop which is not a simple loop, because three nodes have
degree larger than two.

We now define the union of two generalized loops, $l_1 = \langle V_1,
E_1 \rangle$ and $l_2 = \langle V_2, E_2 \rangle$, as the generalized
loop which results from taking the union of the vertices and the edges
of $l_1$ and $l_2$, that is, $l' = l_1 \cup l_2 = \langle V_1 \cup V_2, E_1 \cup E_2 \rangle$.
Note that the union of two simple loops is never a simple loop except
for the trivial case in which both loops are equal.
Figure \ref{fig:ising loop}b shows an example of a generalized loop
which can be described as the union of three simple loops, each
of size 8.
The same example can be also defined as the union of two overlapping
simple loops, each of size 12.

\begin{definition}\label{def:disconnected loop}
A disconnected generalized loop, \textbf{disconnected loop}, is defined as a
generalized loop with more than one connected component.
\end{definition}

Figure \ref{fig:ising loop}c shows an example of a disconnected loop
composed of three simple loops.
Note that components are not restricted to be simple loops.
Figure \ref{fig:ising loop}d illustrates this fact using an example
where one connected component (the left one) is not a simple loop.

\begin{definition}\label{def:complex loop}
A complex generalized loop, \textbf{complex loop}, is defined as a
generalized loop which cannot be expressed as the union of two or more
different simple loops.
\end{definition}

Figures \ref{fig:ising loop}e and \ref{fig:ising loop}f are examples
of complex loops.
Intuitively, they result after the connection of two or more connected
components of a disconnected loop.

Any generalized loop can be categorized according to these three different
categories: a simple loop cannot be a disconnected loop, neither
a complex loop. On the other hand, since Definitions
\ref{def:disconnected loop} and \ref{def:complex loop} are not mutually
exclusive, a disconnected loop can be a complex loop and vice-versa, and
also there are generalized loops which are neither disconnected nor
complex, for instance the example of Figure \ref{fig:ising loop}b.
An example of a disconnected loop which is not a
complex loop is shown in Figure \ref{fig:ising loop}c.
An example of a complex loop which is not a
disconnected loop is shown in Figure \ref{fig:ising loop}e.
Finally, an example of a complex loop which is also a
disconnected loop is shown in Figure \ref{fig:ising loop}f.

\begin{figure}
\begin{center}
\includegraphics[scale=.8]{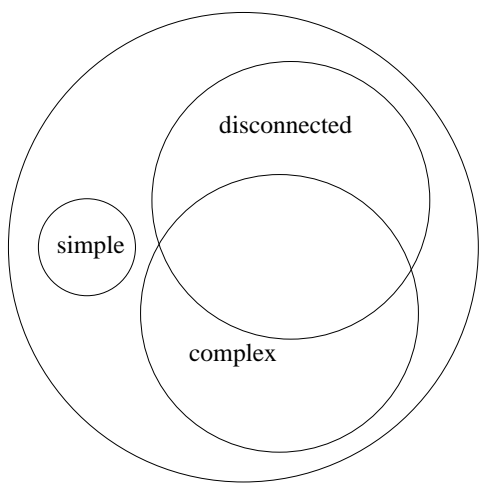}
\end{center}
\caption{
Diagrammatic representation of the different types of generalized
loops present in any graph.
Sizes of the sets are just indicative and depend on the particular
instance. } 
\label{fig:loops-set}
\end{figure}

We finish this characterization using a diagrammatic representation in
Figure \ref{fig:loops-set} which illustrates the definitions.
Usually, the smallest subset contains the simple loops and
both disconnected loops and complex loops have nonempty
intersection. There is another subset of all generalized loops
which are neither simple, disconnected, nor complex.

\section{The Truncated Loop Series Algorithm}
\label{sec:algor}
In this section we describe the TLSBP algorithm to compute generalized
loops in a factor graph, and use them to correct the BP solution.
The algorithm is based on the principle that every generalized loop
can be decomposed in smaller loops.
The general idea is to search first for a subset of the simple loops
and, after that, merge all of them iteratively until no new loops are
produced.
As expected, a brute force search algorithm will only work for small
instances. We therefore prune the search using two different bounds as
input arguments. Eventually, a high number of generalized loops which
presumably will account for the major contributions in the loops
series expansion will be obtained.
We show that the algorithm is complete, or equivalently, that all
generalized loops are obtained by the proposed approach when the
constraints expressed by the two arguments are relaxed.
Although exhaustive enumeration is of little interest for complex
instances, it allows to check the validity of \eqref{eq:series} and to
study the loop series expansion for simpler instances.
The algorithm is composed of three steps:
\begin{enumerate}
\item
First, we remove recursively all the leaves of the original graph,
until its 2-core is obtained.
This initial step has two main advantages.
On the one hand, since some nodes are deleted, the complexity of the
problem is reduced. On the other hand, we can use the resulting
graph as a test for any possible improvement to the BP solution. Indeed,
if the original graph did not contain any loop then the null
graph is obtained, the BP solution is exact on the original graph, and
the series expansion has only one term.
On the other hand, if a nonempty graph remains after this preprocessing,
it will have loops and the BP solution can be improved using the proposed
approach.
\item
After the graph is preprocessed, the second step searches for
simple loops.
The result of this search will be the initial set of loops for the
next step and will also provide a bound $b$ which will be used to
truncate the search for new generalized loops.
Finding circuits in a graph is a problem addressed for
long~\citep{Tiernan70,Tarjan73,Johnson75}
whose computational complexity grows exponentially with the
length of the cycle~\citep{Johnson75}.
Nevertheless, we do not count all the simple loops but only a subset.
Actually, to avoid dependence on particular instances, we parametrize
this search by a size $S$, which limits the number of shortest
simple loops to be considered.
Once $S$ simple loops have been found in order of increasing length,
the length of the largest simple loop is used as the bound $b$ for
the remaining steps.
\item
The third step of the algorithm consists of obtaining all non-simple
loops that the set of $S$ simple loops can``generate''.
\end{enumerate}

According to definition \ref{def:complex loop},
complex loops can not be expressed as union of simple loops.
To develop a complete method, in the sense that all existing loops
can be obtained, we define the operation \emph{merge loops}, which extends the
simple union in such a way that complex loops are retrieved as well.
Given two generalized loops, $l_1, l_2$, \emph{merge loops} returns a
\emph{set} of generalized loops.
One can observe that for each disconnected loop, a set of complex loops
can be generated by connecting two (or more) components of the
disconnected loop.
In other words, complex loops can be expressed as the union of disjoint
loops with a path connecting two vertices of different components.
Therefore the set computed by \emph{merge loops} will have only one
element $l' = \{l_1 \cup l_2\}$ if $l_1 \cup l_2$ is not disconnected.
Otherwise, all the possible complex loops in which $l_1 \cup
l_2$ appears are included in the resulting set.

We use the following procedure to compute all complex loops
associated to the disconnected loop $l'$:
we start at a vertex of a connected component of $l'$ and perform
depth-first-search (DFS) until a vertex of a different
component has been reached. At this point, the connecting path and
the reached component are added to the first component.
Now the generalized loop has one less connected component.
This procedure is repeated again until the resulting generalized loop
is not disconnected, or equivalently, until all its vertices are members of the
first connected component.
Iterating this search for each vertex every time two components are
connected, and also for each initial connected component, one obtains
all the required complex loops.
\floatname{algorithm}{Algorithm}
\begin{algorithm}
\caption{merge loops}
\label{al:mergeLoops}
\begin{algorithmic}[1]
\REQUIRE $\quad$\\
\begin{tabular}[]{l}
$l_1 = \langle V_1, E_1 \rangle$ loop,\\
$l_2 = \langle V_2, E_2 \rangle$ loop,\\
$b$ maximal length of a loop,\\
$M$ maximal depth of complex loops search,\\
$G$ preprocessed factor graph\vspace{.1in}
\end{tabular}
\STATE $newloops \leftarrow \emptyset$
\IF{($|E_1 \cup E_2| \leq b$)}
  \STATE $C \leftarrow \text{Find connected components(}l_1 \cup
  l_2\text{)}$
  \STATE $newloops \leftarrow \{l_1 \cup l_2\}$
  \FORALL{($c_i \in C$)}
    \FORALL{($v_i \in c_i$)}
      \STATE $newloops \leftarrow newloops \cup
      \text{Find complex loopsDFS(}v_i, c_i, C, M, b, G\text{)}$
    \ENDFOR
  \ENDFOR
\ENDIF
\RETURN{$newloops$}
\end{algorithmic}
\end{algorithm}

Note that deciding whether $l_1 \cup l_2$ is disconnected or not
requires finding all connected components of the resulting loop.
Moreover, given a disconnected loop, the number of associated complex loops
can be enormous.
In practice, the bound $b$ obtained previously is used to reduce the
number of calculations.
First, testing if the length of $l_1 \cup l_2$ is larger than $b$ can
be done without computing the connected components.
Second, the DFS search for complex loops is limited using $b$, so very large
complex loops will not be retrieved.

However, restricting the DFS search for complex loops using the bound
$b$ could result in too deep searches.
Consider the worst case of merging the two shortest, non-overlapping,
simple loops which have size $L_s$.
The maximum depth of the DFS search for complex loops is $d = b -
2L_s$.
Then the computational complexity of the merge loops operation depends
exponentially on $d$.
This dependence is especially relevant when $b >> L_s$, for instance
in cases where loops of many different lengths exist.
To overcome this problem we define another
parameter $M$, the maximum depth of the DFS search in the merge loops
operation.
For small values of $M$, the operation \emph{merge loops}
will be fast but a few (if any) complex loops will be obtained.
Conversely, for higher values of $M$ the operation
\emph{merge loops} will find more complex loops at the cost of more time.

Algorithm~\ref{al:mergeLoops} in the previous page 
describes briefly the operation \emph{merge loops}.
It receives two loops $l_1$ and $l_2$, and bounds $b$ and $M$ as
arguments, and returns the set \emph{newloops} which contains the loop
resulting of the union of $l_1$ and $l_2$ plus all complex loops
obtained in the DFS search bounded by $b$ and $M$.
\floatname{algorithm}{Algorithm}
\begin{algorithm}
\caption{Algorithm TLSBP}
\label{al:TLSBP}
\begin{algorithmic}[1]
\REQUIRE $\quad$\\
\begin{tabular}[]{l}
$S$ maximal number of simple loops,\\
$M$ maximal depth of complex loops search,\\
$G$ original factor graph\vspace{.1in}
\end{tabular}
\STATE $\text{Run belief propagation algorithm over }G$
\STATE $G' \leftarrow \text{Obtain the 2-core(}G\text{)}$
\STATE $\mathcal{C}' \leftarrow \emptyset$
\IF{($\neg \text{empty}(G')$)}
  \STATE $\langle sloops, b \rangle \leftarrow \text{Compute first S simple loops(}G'\text{)}$
  \STATE $\langle oldloops, newloops \rangle \leftarrow \langle
  sloops, \emptyset \rangle$
  \STATE $ \mathcal{C}' \leftarrow sloops$
  \WHILE{($\neg \text{empty}(oldloops)$)}
    \FORALL{($l_1 \in sloops$)}
      \FORALL{($l_2 \in oldloops$)}
        \STATE $newloops \leftarrow newloops \cup
        \text{mergeLoops(}l_1, l_2, b, M, G' \text{)}$
      \ENDFOR
    \ENDFOR
    \STATE $oldloops \leftarrow newloops$
    \STATE $\mathcal{C}' \leftarrow \mathcal{C}' \cup newloops$
  \ENDWHILE
\ENDIF
\RETURN{$\text{the result of expression }$\eqref{eq:Zlc} using $\mathcal{C}'$}
\end{algorithmic}
\end{algorithm}

Once the problem of expressing all generalized loops as compositions
of simple loops has been solved using the \emph{merge loops}
operation, we need to define an efficient procedure to merge them.
Note that, given $S$ simple loops, a brute force approach tries
all combinations of two, three, $\hdots$ up to $S-1$ simple loops.
Hence the total number is:
$${S \choose 2} + {S \choose 3} + \hdots + {S \choose S-1} = \mathcal{O}(2^S)$$ 
which is prohibitive.
Nevertheless, we can avoid redundant combinations by merging pairs of
loops iteratively:
in a first iteration, all pairs of simple loops are merged, which
produces new generalized loops.
In a next iteration $i$, instead of performing all ${S \choose i}$
mergings, only the new generalized loops obtained in 
iteration $i-1$ are merged with the initial set of simple loops.
The process ends when no new loops are found.
Using this merging procedure, although the asymptotic cost is still exponential
in $S$, many redundant mergings are not considered. 

Summarizing, the third step applies iteratively the 
\emph{merge loops} operation until no new generalized loops are
obtained.
After this step has finished, the final step computes the truncated
loop corrected partition function defined in Equation~\eqref{eq:Zlc} using
all the obtained generalized loops.
We describe the full procedure in Algorithm~\ref{al:TLSBP}.
Lines $2$ and $4$ correspond to the first and second steps and lines
$5-13$ correspond to the third step.

To show that this process produces all the generalized loops we first
assume that $S$ is sufficiently large to account for all the
simple loops in the graph, and that $M$ is larger or equal than the number
of edges of the graph.
Now let $C$ be a generalized loop.
According to the definitions of Section~\ref{sec:charac}, either $C$
can be expressed as a union of $s$ simple loops, or $C$ is a
complex loop.
In the first case, $C$ is clearly produced in the $s$th iteration.
In the second case, let $s'$ denote the number of simple loops which
appear in $C$.
Then $C$ is produced in iteration $s'$, during the DFS for
complex loops within the merging of one of the simple loops contained
in $C$.

The obtained collection of loops can be used for the approximation of
the singe node marginals as well, as described in Equation~\eqref{eq:Zclamp-tlsbp}.
The method consists of clamping one variable $i$ to all its possible values ($\pm 1$)
and computing the corresponding approximations of the partition functions:
$Z_{TLSBP}^{x_i=+1}$ and $Z_{TLSBP}^{x_i=-1}$.
This requires to run BP in each clamped network, and reuse the set of loops
replacing with zero those terms where the clamped variable appears.
The computational complexity of approximating all marginals using this approach
is in general $\mathcal{O}(N\cdot L\cdot d\cdot T_{BP})$, where $L$ is the
number of found loops, $d$ is the cardinality of the variables (two in our case),
and $T_{BP}$ the average time of BP to converge after clamping one variable.
Usually, this task requires less computation time than the search for loops.

As a final remark, we want to stress a more technical aspect related to
the implementation.
Note that generalized loops can be expressed as the composition of
other loops in many different ways.
In consequence, they all must be stored incrementally and the operation of
checking if a loop has been previously counted or not should be done
efficiently.
An appropriate way to implement this fast look-up/insertion is to
encode all loops in a string composed by the edge identifiers in some
order with a separator character between them.
This identifier is used as a key to index an ordered tree, or hash
structure.
In practice, a hash structure is only necessary if large amounts of
loops need to be stored.
For the cases analyzed here, choosing a balanced tree instead of a hash
table resulted in a more efficient data structure.


\section{Experiments}
\label{sec:experim}
In this section we show the performance of TLSBP in two different
scenarios.
First, we focus on square lattices and study how loop
corrections improve the BP solution as a function of the interaction
between variables and the size of the problem.
Second, we study the performance of the method in random regular
graphs as a function of the degree between the nodes.

In all the experiments we show results for tractable instances,
where the exact solution using the junction tree~\citep{lauritzen88} can
be computed.
Performance is evaluated comparing the TLSBP error
against the BP solution, and also against the cluster variation method (CVM).
Instead of using a generalized belief propagation algorithm (GBP)
which usually requires several trials to find the proper damping factor
to converge, we use a double-loop implementation which has convergence guarantees
~\citep{hes03a}.
For this study we select as outer regions of the CVM method all maximal factors
together with all loops that consist up to four different variables.
This choice
represents a good trade-off between computation time required for convergence
and accuracy of the solution.

We report two different error measures. Concerning the partition function
$Z$ we compute:
\begin{align}\label{eq:belerror}
\text{Error}_{Z'} & =   \left |\frac{\log Z'}{\log Z}\right |,
\end{align}
where $Z'$ is the partition function corresponding to the method used: BP, TLSBP,
or CVM.
Error of single-node marginals was measured using the maximum $\ell_\infty$ error,
which is a reasonable quantity if one is interested in worst-case scenarios:
\begin{align}\label{eq:logzerror}
\text{Error}_b & = \max_{\scriptstyle \substack{i=1,\ldots,n \\ x_i={\pm 1}}} {\lvert{P_i(x_i) - b_i(x_i)}}\rvert ,
\end{align}
were again $b_i(x_i)$ are the single-node marginal approximations corresponding
to the method used.

We use four different schemas for belief-updating of BP: \emph{(i)} fixed
and \emph{(ii)} random sequential updates, \emph{(iii)} parallel (or synchronous) updates,
and \emph{(iv)} residual belief propagation (RBP), a recent method proposed in \citep{Elidanal}.
The latter method schedules the updates of the BP messages heuristically
by selecting the next message to be updated which has maximum \emph{residual},
a quantity defined as an upper bound on the distance of the current messages
from the fixed point.
In general, we experienced that for some instances where the RBP method converged,
the other update schemas (fixed, random sequential and parallel updates) failed
to converge.

In all schemas we interpret that a fixed point is reached at iteration $t$
when the maximum absolute value of the updates of all the messages from iteration $t-1$ to $t$
is smaller than a threshold $\vartheta$.
We notice a large correlation between the order of magnitude of $\vartheta$ and the
ratio between the BP and the TLSBP errors.
For this reason we used a very small value of the threshold, $\vartheta = 10^{-17}$.

\subsection{Ising Grids}
\label{subsec:ising1}
\begin{figure}[!b]
\begin{center}
\includegraphics[width=.45\linewidth]{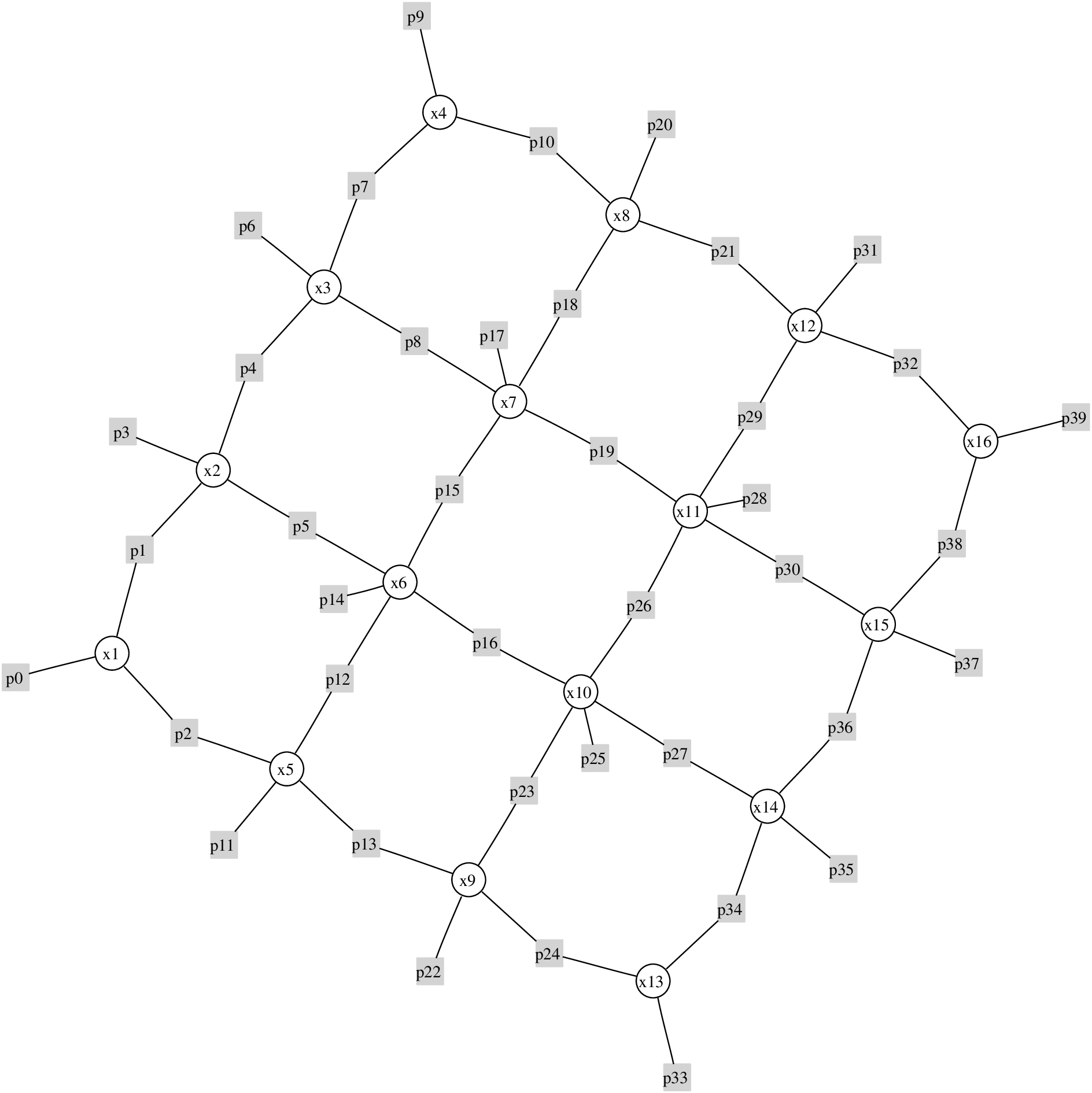} 
\includegraphics[height=7cm,width=.45\linewidth]{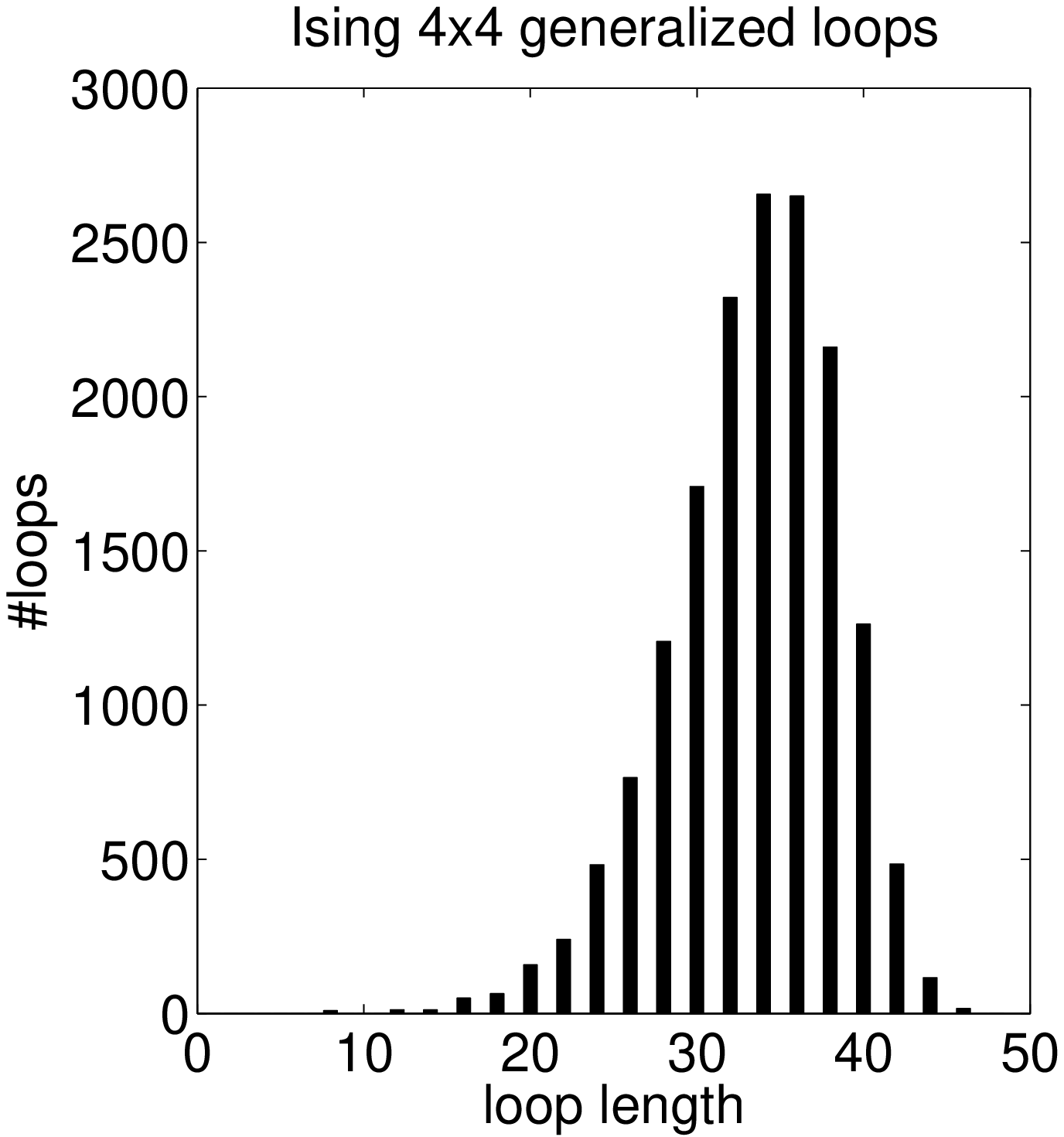} 
\end{center}
\caption{\textbf{(left)} A factor graph representing the 4x4 Ising
  grid. \textbf{(right)} Number of generalized loops as a function of the length
  using the factor graph representation.}
\label{fig:ising4x4}
\end{figure}

This model is defined on a grid where each variable,
also called spin, takes binary values $x_i = \pm 1$.
A spin is coupled with its direct neighbors only, so that pairwise
interactions $f_{ij}(x_i,x_j) = \exp (\theta_{ij} x_i x_j)$ are considered, 
parametrized by $\theta_{ij}$.
Every spin can be exposed to an external field $f_i(x_i) = \exp (\theta_i x_i)$, 
or single-node potential, parametrized by $\theta_i$. 
Figure~\ref{fig:ising4x4} (left) shows the factor graph associated to the 4x4 Ising
grid, composed of 16 variables.
The Ising grid model is often used as a test-bed for inference algorithms.
It is of great relevance in statistical physics, and has applications in different
areas such as image processing.
In our context it also represents a challenge since it has
many loops.
Good results in this model will likely translate into good results for
less loopy graphs.

Usually, two cases are differentiated according to the sign of the $\theta_{ij}$
parameters. For $\theta_{ij} > 0$ coupled spins tend to be in the same state.
This is known as the attractive, or ``ferromagnetic'' setting.
On the other hand, for mixed interactions, $\theta_{ij}$ can be
either positive or negative, and this setting is called ``spin-glass'' configuration.
Concerning the external field, one can distinguish two cases.
For the case of nonzero fields, larger values of $\theta_i$ imply easier
inference problems in general.
On the other hand, for $\theta_i = 0$, there exist two
phase transitions from easy inference problems (small $\theta_{ij}$) to more
difficult ones (large $\theta_{ij}$) depending on the type of pairwise
couplings~\citep[see][for more details]{mooij2005}.

This experimental subsection is structured in three parts:
First, we study a small 4x4 grid.
We then study the performance of the algorithm in a 10x10 grid, where
complete enumeration of all generalized loops is not feasible.
Finally, we analyze the scalability of the method with problem size.

\label{page-ref}
The 4x4 Ising grid is complex enough to account for all types of
generalized loops.
It is the smallest size where complex loops are present.
At the same time, the problem is still tractable and exhaustive
enumeration of all the loops can be done.

We ran the TLSBP algorithm in this model with arguments $S$ and $M$
large enough to retrieve all the loops.
Also, the maximum length $b$ was constrained to be $48$, the total number
of edges for this model.
After 4 iterations all generalized loops were obtained.
The total number is $16371$ from which $213$ are simple loops.
The rest of generalized loops are classified as follows:
$174$ complex and disconnected loops,
$1646$ complex but non-disconnected loops,
$604$ non-complex but disconnected loops,
and $13734$ neither complex nor disconnected loops.

Figure~\ref{fig:ising4x4} (right) shows the histogram of all generalized loops for
this small grid. Since we use the factor graph representation
the smallest loop has length 8. The largest generalized loop includes
all nodes and all edges of the \emph{preprocessed} graph, and has
length 48.
The Poisson-like shape of the histogram is a characteristic of this
model and for larger instances we observed the same tendency.
Thus the analysis for this small model can be extrapolated to
some extent to grids with more variables.

To analyze how the error changes as more loops are considered it is
useful to sort all the terms $r(C)$ by their absolute value in descending
order such that $\left|r(C_i)\right| \geq \left|r(\mathcal{C}_{i+1})\right|$.
We then compute, for each number of loops $l = 1 \ldots 16371$, the approximated partition
function which accounts for the $l$ most important loops:
\begin{align}\label{eq:cum}
Z_{TLSBP}(l) = & Z_{BP} \left( 1 + \sum_{i=1...l}{r(C_i)} \right),
\end{align}
\begin{figure}[!t]
\begin{center}
\includegraphics[scale=.54]{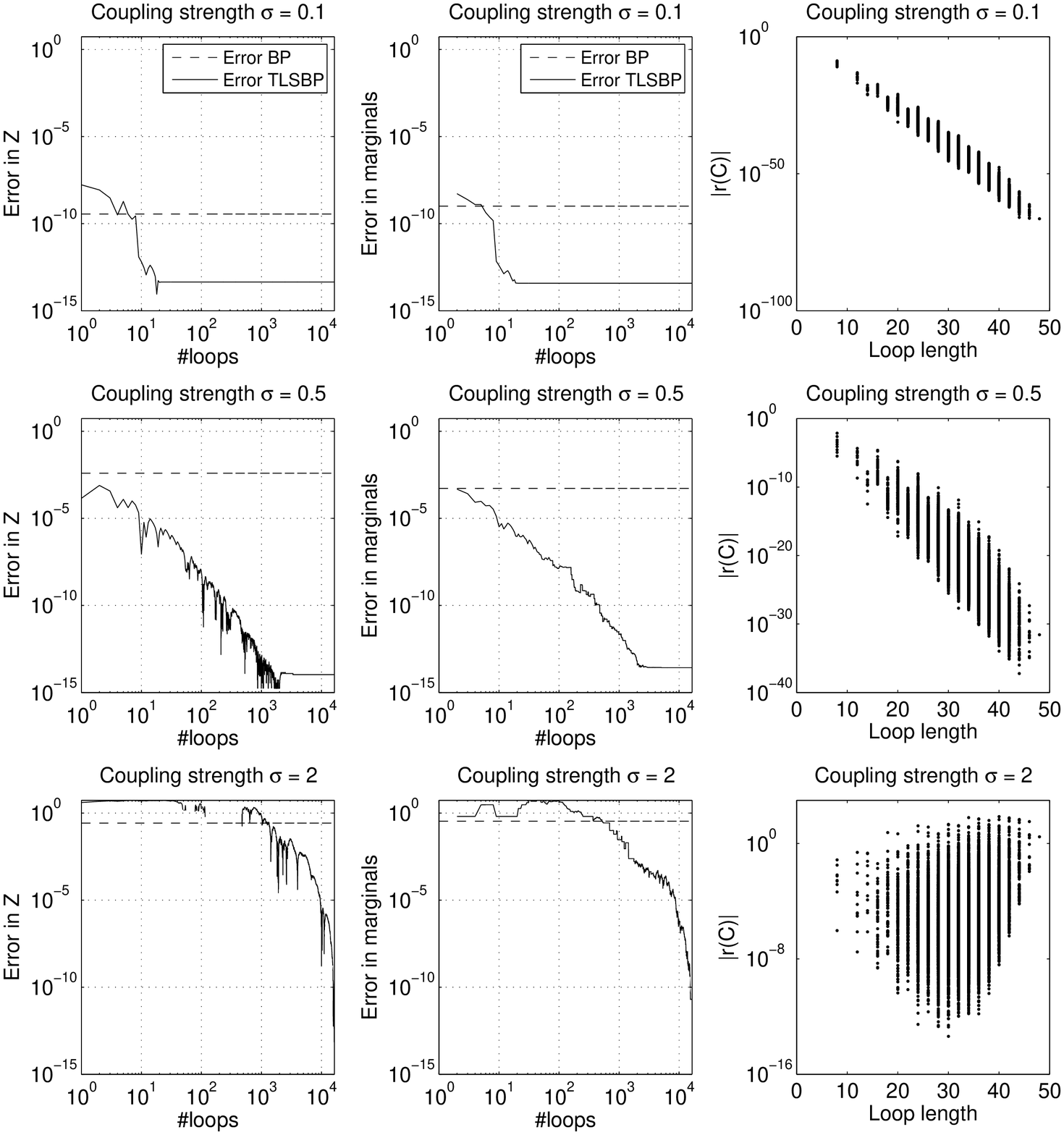} 
\end{center}
\caption{Cumulative error for the spin-glass
  4x4 Ising grid for different interaction strengths, see Equation~\eqref{eq:cum}.
  \textbf{(left column)} Error of Z.
  \textbf{(middle column)} Error of single-node marginals.
  Dashed lines correspond to the BP error, and solid lines correspond to
  the loop-corrected (TLSBP) error.
  \textbf{(right column)} Absolute values of all loop terms as a function
  of the length of the corresponding loop.
}
\label{fig:ising4x4_coupling}
\end{figure}

From these values of the partition function we calculate the error measure
indicated in Equation~\eqref{eq:logzerror}.
Estimations of the single-node marginals were obtained using the clamping
method, and their corresponding error was calculated using Equation~\eqref{eq:belerror}.

We now study how loop contributions change as a function of the
coupling strength between the variables.
We ran several experiments using mixed
interactions with $\theta_{ij} \sim \mathcal N(0, \sigma^2)$
independently for each factor node, and $\sigma$ varying between $0.1$
and $2$.
Single-node potentials were drawn according to $\theta_i \sim
\mathcal N(0,0.05^2)$.
For small values of $\sigma$, interactions are weak and BP converges easily,
whereas for high values of $\sigma$ variables are strongly coupled and BP has
more difficulties, or does not converge at all.

Figure~\ref{fig:ising4x4_coupling} shows results of 
representative instances of three different interaction strengths.
For each instance we plot the partition function error (left column)
together with errors in the single-node marginals (middle column) and
loop contributions as a function of the length (right column).
First, we can see that improvements of the partition sum correspond to
improvements of the estimates of marginal probabilities as well.
Second, for weak couplings ($\sigma = 0.1$, first row) we
can see that truncating the series until a small number of loops (around
$10$) is enough to achieve machine precision.
In this case the errors in BP are most prominently due to small
simple loops. As the right column illustrates, loop contributions decrease
exponentially with the size, and loops with the same length correspond to
very similar contributions.
Larger loops give negligible contributions and can thus be ignored by
truncating the series.
As interactions are strengthened, however, more loops have to be
considered to achieve maximum accuracy, and contributions show more
variability for a given length (see middle row).
Also, oscillations of the error due to the different signs in loop terms (caused by
the mixed interactions) of the same order of magnitude become more frequent.
Eventually, for large couplings ($\sigma \geq 2$), where BP often fails to
converge, loops of all lengths give significant contributions.
In the bottom panels of Figure~\ref{fig:ising4x4_coupling} we show an
example of a 'difficult' case for which the BP result is not improved
until more than $10^3$ loop terms are summed.
The observed discontinuities in the error of the partition sum are
caused by the fact that oscillations become more pronounced, and
corrections composed of negative terms $r_i(C_i)$ can result in
negative values of the partially corrected partition function,
see Equation~\eqref{eq:cum}.
This occurs for very strong interactions only, and when a small fraction
of the total number of loops is considered.
In addition, as the right column indicates, there is a shift of the
main contributions towards the largest loops.

After analyzing a small grid, we now address the case of the 10x10 Ising grid,
where exhaustive enumeration of all the loops is not computationally feasible.
We test the algorithm in two scenarios: for attractive interactions
(ferromagnetic model) where pairwise interactions are parametrized as
$\theta_{ij} = |\theta'_{ij}|, \theta'_{ij} \sim \mathcal N(0, \sigma^2)$,
and also for the previous case of mixed interactions
(spin-glass model). Single-node potentials were chosen $\theta_{i} \sim \mathcal N(0.1, 0.05^2)$ 
in both cases.

\begin{figure}[!t]
\begin{center}
\includegraphics[scale=.51]{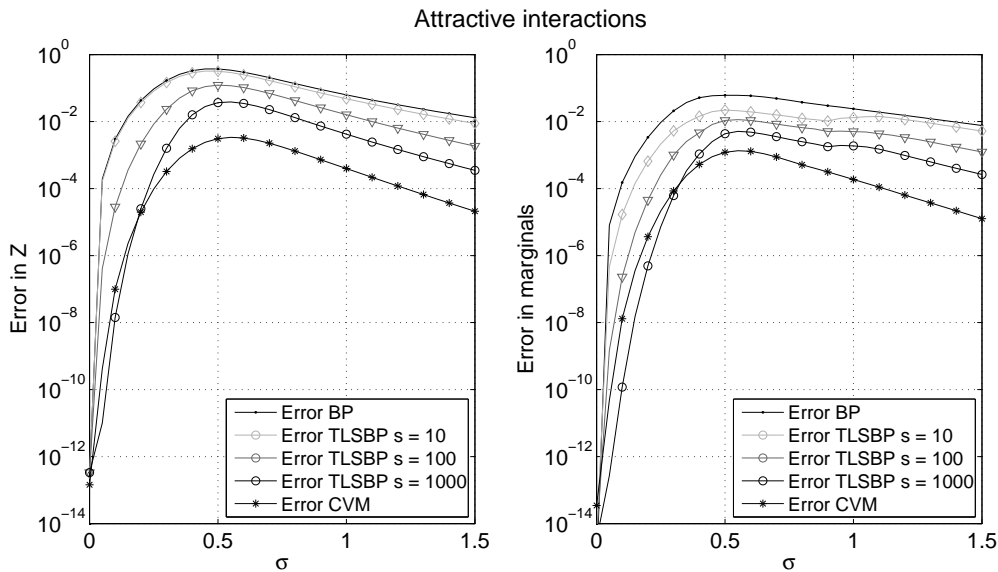}
\end{center}
\caption{TLSBP error for the 10x10 Ising grid with attractive 
interactions for different values of the parameter $S$.
\textbf{(left)} Error of the partition
function. \textbf{(right)} Error of single-node marginals.}
\label{fig:ising10x10_coupling_fg}
\end{figure}

\begin{figure}[!t]
\begin{center}
\includegraphics[scale=.51]{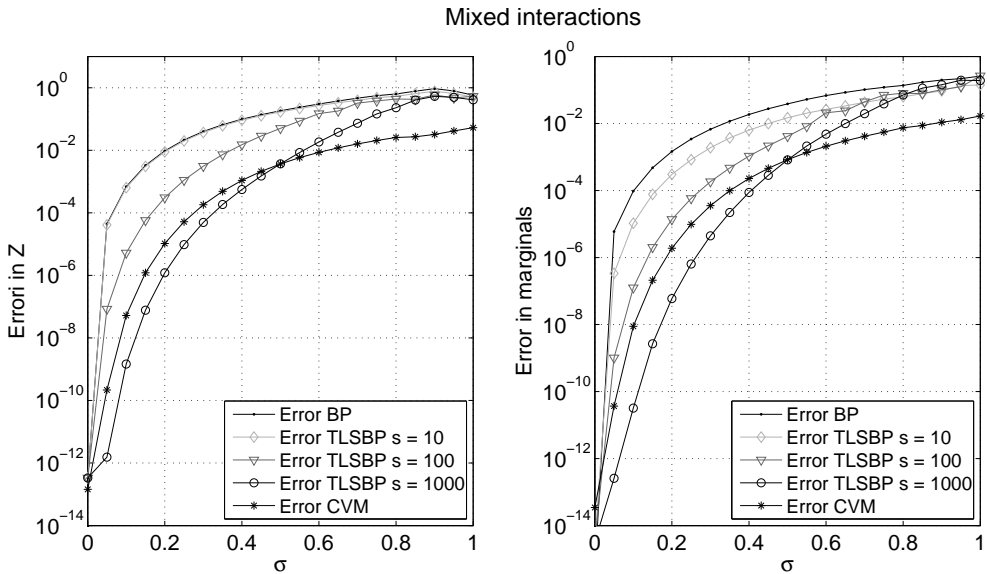}
\end{center}
\caption{TLSBP error of the 10x10 Ising grid with mixed 
interactions for different values of the parameter $S$.
\textbf{(left)} Error of the partition
function. \textbf{(right)} Error of single-node marginals.}
\label{fig:ising10x10_coupling_sg}
\end{figure}


We show results in Figures~\ref{fig:ising10x10_coupling_fg} and
\ref{fig:ising10x10_coupling_sg} for three
values of the parameter $S = \{10, 100, 1000\}$ and a fixed value of
$M=10$.
For $S=10$ and $S=100$ only simple loops were obtained whereas for
$S=1000$ a total of $44590$ generalized loops was used to compute
the truncated partition sum.
Results are averaged errors over $50$ random instances.
The selected loops were the same in all instances.
Although in both types of interactions the BP error (solid line with dots) is
progressively reduced as more loops are considered, the picture
differs significantly between the two cases.

For the ferromagnetic case shown in Figure~\ref{fig:ising10x10_coupling_fg}
we noticed that all loops have positive contributions, $r(C) > 0$.
This is a consequence of this particular type of interactions,
since all magnetizations have the same sign at the BP fixed point, and also
all nodes have an even number of neighbors.
Consequently, improvements in the BP result are monotonic as more loops are considered,
and in this case, the TLSBP can be considered as a lower bound of the exact
solution.
For the case of $S = 1000$, the BP error is reduced substantially
at nonzero $\sigma$, but around $\sigma \sim
0.5$, where the BP error reaches a maximum, also the TLSBP improvement
is minimal.
From this maximum, the BP error decreases again, and loop
corrections tend to improve progressively the BP solution again as the coupling is
strengthened.
We remark that improvements were obtained for all instances in the three
cases.

Comparing with CVM, TLSBP is better for weak couplings and for $S=1000$ only.
This indicates that for intermediate and strong couplings one would need
more than the selected $44590$ generalized loops to improve on the CVM
result.

For the case of spin-glass interactions we report different behavior.
From Figure~\ref{fig:ising10x10_coupling_sg} we see again
that for weak couplings the BP error is corrected substantially,
but the improvement decreases as the coupling strength is increased.
Eventually, 
for $\sigma \sim 1$ BP fails to converge in most of the cases and
also gives poor results.
In these cases loop corrections are of little use, and there is no actual
difference in considering $S=1000$ or $S=10$.
Moreover, because loop terms $r(C)$ now can have different signs, truncating the
series can lead to worse results for $S=1000$ than for $S=10$.
Interestingly, the range where TLSBP performs better than CVM is slightly larger
in this type of interactions, TLSBP being better for $\sigma < 0.5$.

To end this subsection, we study how loop corrections scale with the
number of nodes in the graph.
We only use spin-glass interactions, since it is a more difficult
configuration than the ferromagnetic case, as previous experiments
suggest.
We compare the performance for weak couplings ($\sigma = 0.1$), and strong
couplings ($\sigma = 0.5$), where BP has difficulties to converge in
large instances.
The number of variables $N^2$ is increased for grids of size $N\times N$
until exact computation using the junction tree algorithm is not
feasible anymore.

\begin{figure}[!t]
\begin{center}
\includegraphics[scale=.5]{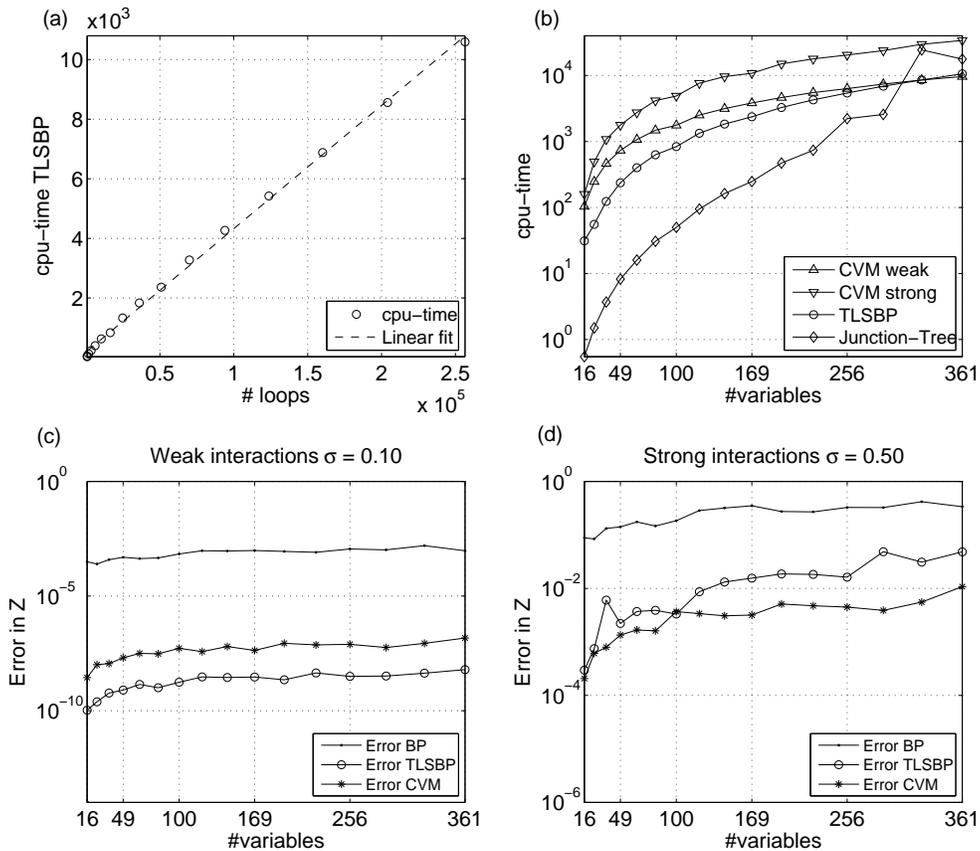}
\end{center}
\caption{Scalability of the method in the Ising model.
\textbf{(a)} Time complexity as a function of the produced number
of generalized loops. \textbf{(b)} Relation between the time
complexity of TLSBP and CVM.
Comparison of the error of the partition function between BP, TLSBP and CVM
as a function of the graph size for \textbf{(c)} weak interactions
and \textbf{(d)} strong interactions.
}
\label{fig:ising_scaling}
\end{figure}
Since the number of generalized loops grows very fast with the size of
the grid, we choose increasing values of $S$ as well.
We use values of $S$ proportional to the number of variable nodes $N^2$ such that
$S = 10 N^2$.
This simple linear increment in $S$ means that as $N$ is increased, the
proportion of simple loops captured by TLSBP over the total existing number of
simple loops decreases.
It is interesting to see how this affects the performance of TLSBP in
terms of time complexity and accuracy of the solution.
For simplicity, $M$ is fixed to zero, so no complex loops are considered.
Moreover, to facilitate the computational cost comparison, we only
compute mergings of pairs of simple loops.
Actually, for large instances the latter choice does not modify the
final set of loops, since generalized loops which can only be expressed as
compositions of three or more simple loops are pruned using the bound
$b$.

In Figure~\ref{fig:ising_scaling}, the top panels show averaged
results of the computational cost.
The left plot indicates the relation between the number of loops
computed by TLSBP and the time required to compute them.
This relation can be fit accurately using a line which means that for this
choice of parameters $S$ and $M$, and considering only mergings of simple loops,
the computational complexity of the algorithm grows just linearly with the found loops.
One has to keep in mind that the number of loops obtained using the
TLSBP algorithm grows much faster, but much less than the total number of existing
loops in the model.

Figure~\ref{fig:ising_scaling}b shows the CPU time consumed by CVM, TLSBP, and
the junction tree algorithm.
In this case, since we only compute the partition function $Z$, the CPU time of TLSBP
is constant for both weak and strong couplings.
On the contrary, CVM depends on the type of interactions.
For weak interactions, TLSBP is in general more efficient than CVM, although the scaling
trend is slightly better for CVM. For $N=19$, CVM starts to be more efficient than TLSBP.
For strong interactions, CVM needs significantly more time to converge in all cases.
If we compare the computational cost of the exact method against TLSBP,
we can see that the junction tree is very efficient for networks with small $N$, and the
best option in those cases.
However, for $N>17$, the junction tree needs more computation time, and
for $N>19$, the tree-width of the resulting grids is too large.
TLSBP memory requirements were considerably less in these cases,
since loops can be stored efficiently using sets of chars.
Also, we can see that the TLSBP scaling is better for this choice of parameters than
that of the exact method.


Bottom panels show the accuracy of the TLSBP solution.
For weak couplings (bottom-left) the BP error is always decreased significantly
for this choice of parameters and the improvement remains almost constant as $N$
increases, meaning that, in this case the number of loops which 
contribute most to the series expansion does not grow significantly with
$N$.
Interestingly, results are always better than CVM for this regime.

For strong couplings (bottom-right) the picture changes.
First, results differ more between instances causing a less smooth
curve.
Second, the TLSBP error also increases with the problem size, so
improvements tend to decrease with $N$, even faster than the BP error decay.
Eventually, for the largest tractable instance the TLSBP improvement is still
significant, about one order of magnitude.
Comparing against CVM, unlike in the weak coupling scenario, the TLSBP method
does not seem to perform better, and only for some cases TLSBP error
is comparable to the CVM error on average.
The accuracy of the TLSBP solution for these instances can be increased by
considering larger values of $S$ and $M$, at the cost of more time.

\subsection{Random Graphs}
\label{sec:rndgraphs}
The previous experimental results were focused on the Ising grid 
which only considers pairwise and singleton interactions in such a way
that each node in the graph is at most linked with four neighbors.
Here we briefly analyze the performance of TLSBP applied on a more general
case, where interactions are less restricted.

We perform experiments on random graphs with regular topology, where
each variable is coupled randomly with $d$ other variables using
pairwise interactions parametrized by
$\theta_{ij} \sim \mathcal{N}(0, \sigma^2)$.
Single-node potentials were parametrized in this case by
$\theta_{i} \sim \mathcal{N}(0, 0.05^2)$.
We study how loop corrections improve the BP solution as a function of
the degree $d$, and compare improvements against the CVM.
As in the previous subsection, for CVM we select the loops of four variables
and all maximal factors as outer clusters.

Note that the rate of increase in the number of loops with the degree
$d$ is even higher than with the number of variables in the Ising
model.
Adding one more link to all the variables means adding $N$ more factor
nodes to the factor graph. This raises the number of
loops dramatically.

\begin{figure}[!tr]
\begin{center}
\includegraphics[scale=.52]{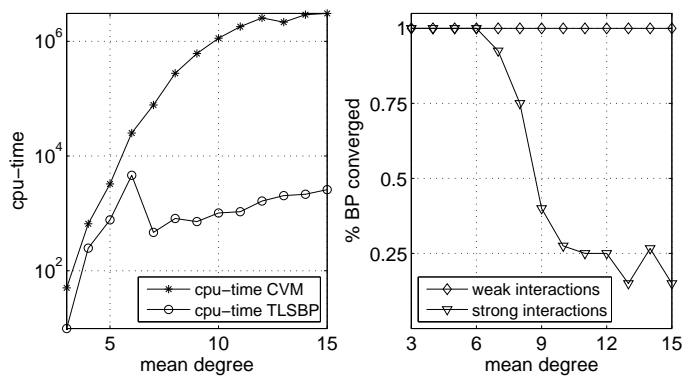}
\end{center}
\caption{
Results on random regular graphs. 
\textbf{(a)} Computation time of TLSBP and CVM. In this case,
we averaged also all instances over all $60$ weak and $60$ strong interactions,
since costs were very similar in both cases.
\textbf{(b)} Fraction of the instances were BP converged.
No convergence is reported when none of the four proposed schedules converged.
}
\label{fig:random_degree_cpu}
\end{figure}

For this scenario, we use $N = 20$ variables and also increase $S$
every time $d$ is increased. We simply start with $S=10$ and use
increments of $250$ for each new $d$.
$M$ was set to $10$, and all possible mergings were computed.
We analyze two scenarios, weak ($\sigma = 0.1$) and strong
couplings ($\sigma = 0.5$), and report averages over $60$ random instances
for each configuration.
As Figure~\ref{fig:random_degree_cpu} (right) indicates, for $\sigma = 0.1$
BP converged in all instances, whereas for $\sigma = 0.5$ BP convergence becomes 
more difficult as we increase $d$.
\begin{figure}[!b]
\begin{center}
\includegraphics[scale=.52]{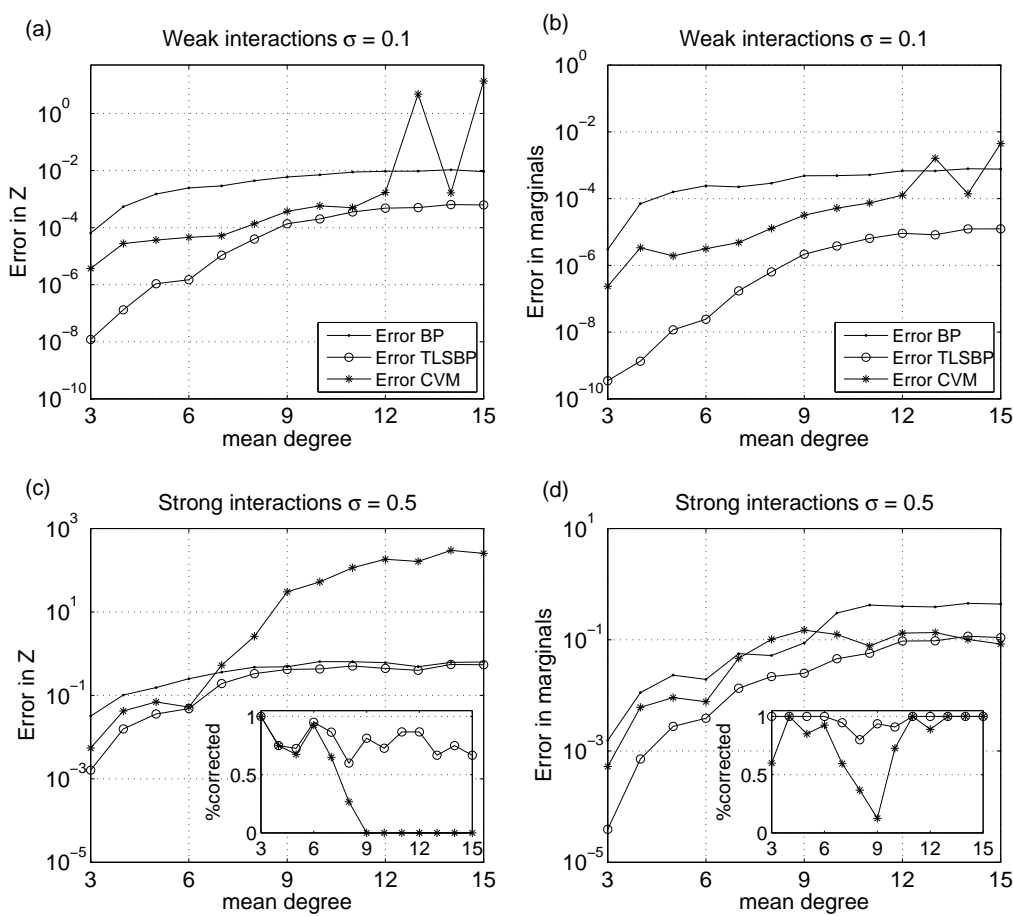}
\end{center}
\caption{Results on random regular graphs. TLSBP and CVM errors as a function
  of the degree $d$. Results are averages over $60$ random instances.
  Errors in the partition function for weak interactions \textbf{(a)},
  marginals for weak interactions \textbf{(b)},
  partition function for strong interactions \textbf{(c)}, and
  marginals for strong interactions \textbf{(d)}.
  Insets show percentage of instances where the BP error was corrected.
}
\label{fig:random_degree_error}
\end{figure}

Figure~\ref{fig:random_degree_error} (top) shows results for weak interactions.
The TLSBP algorithm always corrects the BP
error, although as $d$ increases, the improvement is progressively reduced.
We also notice that in all cases and methods the approximation of the partition
function (left) is less accurate than the approximation of the marginals (right).
For $d=15$, TLSBP improvements are still about one order of
magnitude for the partition function, and even better for the marginals.
As in previous experiments with square lattices, the TLSBP approach is generally
better than CVM in the weak coupling regime.
Here, it is also more stable, since for some dense networks the CVM error can be 
very large, as we can see for $d=13$ and $d=15$.

For strong interactions (bottom panels), we see that differences between
approximations of the partition function and single-node marginals are more
remarkable than in the previous case.
The BP partition function is corrected by TLSBP in more than half of the instances for all
degrees (see inset of Figure~\ref{fig:random_degree_error}c, where we plot the
fraction of instances where BP was corrected in those cases that converged),
although for higher degrees, the TLSBP corrections are small
using this choice of parameters.
On the other hand, single-node BP marginals are corrected in almost all cases. 
In contrast, the CVM approach with our selection of outer clusters does not perform
better than TLSBP in general.
In particular, we see that CVM estimates of the partition function are
very degraded as networks become more dense.
This unsatisfactory performance of CVM in the estimation of the partition function 
is not as noticeable in the marginal estimates, where BP results are often improved,
although with much more variability than the TLSBP method.
Interestingly, for those few instances of dense networks for which BP converged,
CVM estimates of the marginals were very similar to TLSBP.

Finally, we compare computational costs in Figure~\ref{fig:random_degree_cpu} (left).
CVM requires significantly more time to converge than the time required by TLSBP
searching for loops and calculating marginals.
If we analyze in detail how the TLSBP cost changes, we can notice different
types of growth for $d<7$ and for $d \geq 7$.
The reason behind these two scaling tendencies can be explained by the 
choice of TLSBP parameters, and the bound $b$ (the size of the largest simple loop).
For $d<7$, many simple loops of different lengths are obtained.
Consequently, the cost of the merging step grows fast, since many loops with
length smaller than $b$ are produced.
On the other hand, for $d \geq 7$ simple loops have similar lengths and,
therefore, less combinations result in additional loops with length larger
than the bound $b$.
Without bounding the length of the loops in the merging step,
we would expect the first scaling tendency ($d<7$) also for values of $d \geq 7$.

From these experiments we can conclude that TLSBP performance is generally better
than CVM in this domain.
We should mention that alternative choices of regions would have
lead to different CVM results, but will probably not change this conclusion.

\subsection{Medical Diagnosis}
\label{sec:promedas}
We now study the performance of TLSBP on a
``real-world'' example, the Promedas medical diagnostic network.
The diagnostic model in Promedas is based on a Bayesian network.
The global architecture of this network is similar to QMR-DT 
\citep{shwe91}.
It consists of a diagnosis layer that is connected to a layer with 
findings.
In addition, there is a layer of variables, such as age and gender,
that may affect the prior probabilities of the diagnoses.
Since these variables are always clamped for each patient case,
they merely change the prior disease probabilities and are irrelevant
for our current considerations.
Diagnoses (diseases) are modeled as a priori independent binary variables
causing a set of symptoms (findings) which constitute the bottom layer. 
The Promedas network currently consists of
approximately 2000 diagnoses and 1000 findings.

The interaction between diagnoses and findings is modeled with a noisy-OR
structure.  The conditional probability of the finding given the parents
is modeled by $n+1$ numbers, $n$ of which represent the probabilities that
the finding is caused by one of the diseases and one that the finding
is not caused by any of the parents. 

The noisy-OR conditional probability tables with $n$ parents can be
naively stored in a table of size $2^n$. This is problematic for the
Promedas networks since findings that are affected by more than 30
diseases are not uncommon. 
We use
efficient implementation of noisy-OR relations as proposed by
\citep{ambrosio99multiplicative} to reduce the size of these tables.
The trick
is to introduce dummy variables $s$ and to make use of the property
\begin{eqnarray}
{\rm OR}(x|y_1,y_2,y_3) = \sum_{s}{\rm OR}(x|y_1,s){\rm OR}(s|y_2,y_3)
\end{eqnarray}
The interaction potentials on
the right hand side involve at most three variables instead of the initial four
(left). Repeated application of this formula reduces all tables to three
interactions maximally.

\begin{figure}
\begin{center}
\includegraphics[height=7cm,width=6.5cm]{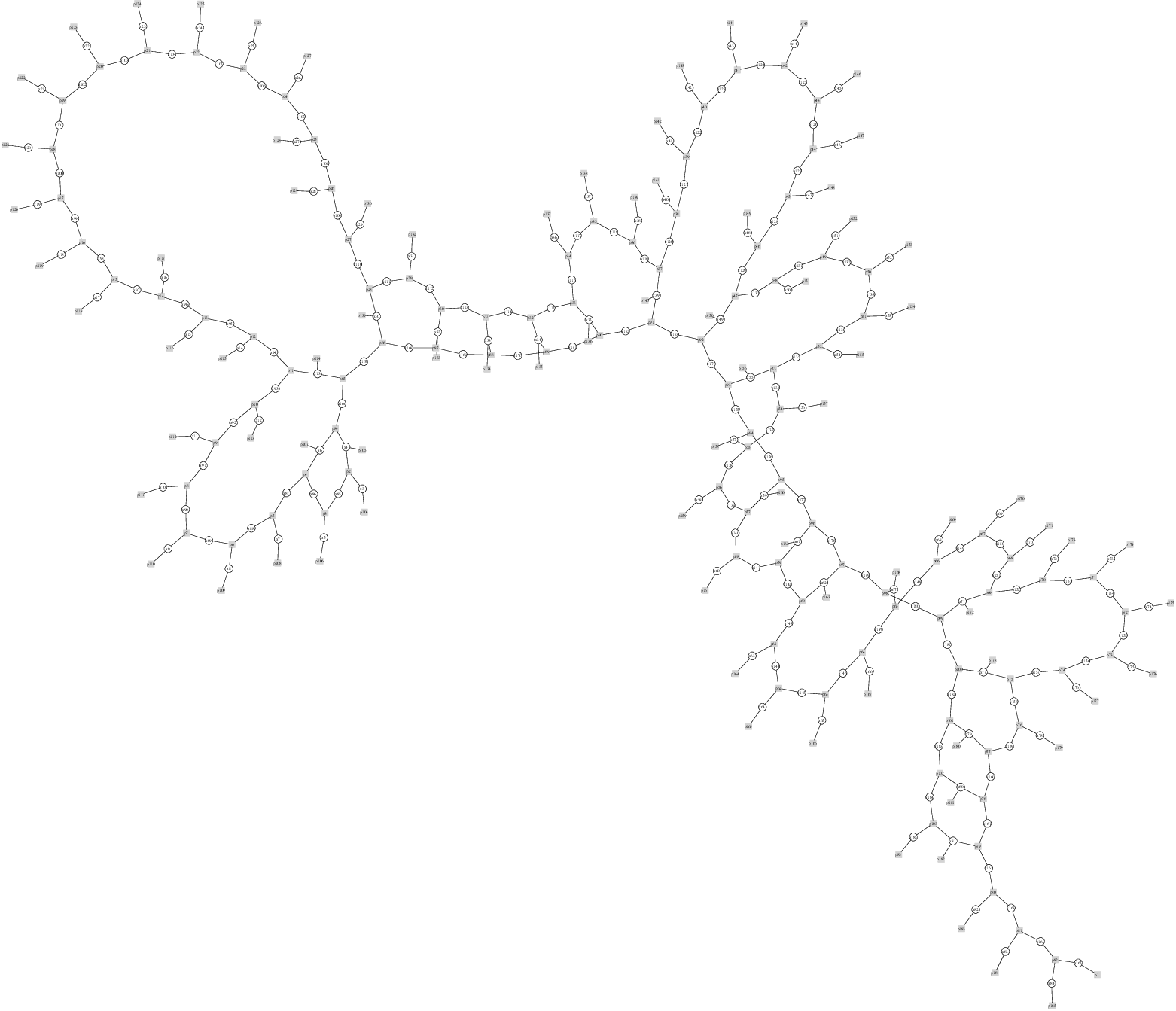} 
\includegraphics[height=7cm,width=5cm]{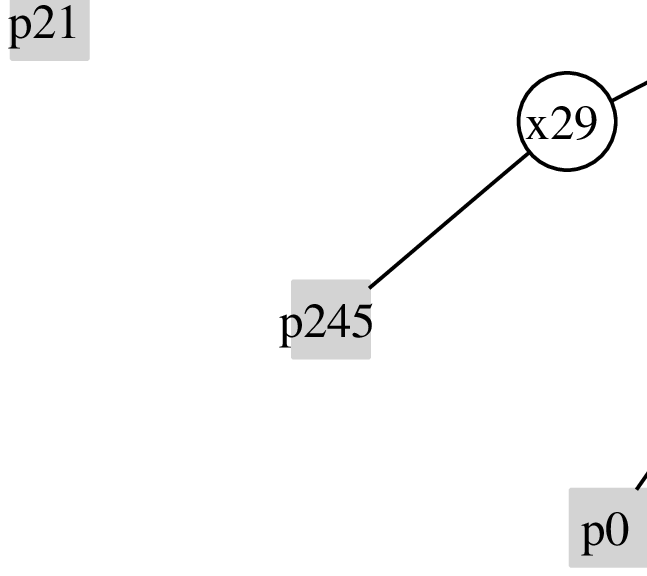}
\end{center}
\caption{
Examples of graph structures, corresponding to patient cases generated
with one disease, after removal of unclamped findings and irrelevant
disease variables and the introduction of dummy variables.
Left and right graphs corresponds to an ``easy'' and a ``difficult''
case respectively.}
\label{fig:typicalgraphs}
\end{figure}

When a patient case is presented to Promedas, a subset of the findings
will be clamped and the rest will be unclamped. If our goal is to
compute the marginal probabilities of the diagnostic variables only, 
the unclamped findings and the diagnoses that are not related to any
of the clamped findings can be summed out of the network as a
preprocessing step. 
The clamped findings cause an effective interaction between
their parents. 
However, the noisy-OR structure is such that when the
finding is clamped to a negative value, the effective interaction
factorizes over its parents. Thus, findings can be clamped to negative
values without additional computation cost \citep{jaakkola_jordan99}.

The complexity of the problem now depends on the set of findings that
is given as input. 
The more findings are clamped to a positive value, the larger
the remaining network of disease variables and the more complex the
inference task.
Especially in cases where findings share more than
one common possible diagnosis, and consequently loops occur, the model
can become complex.  We illustrate some of the graphs that result
after pruning of unclamped
findings and irrelevant diseases and the introduction of dummy variables
for some patient cases in Figure~\ref{fig:typicalgraphs}.

We use the Promedas model to generate virtual patient data
by first clamping one disease variable to a positive value and then
clamping a finding to its positive value with probability equal to the
conditional distribution of the findings given this positive disease.
The union of all positive findings thus obtained constitute one patient
case. For each patient case, the corresponding truncated graphical model
is generated.  Note that the number of disease nodes in this graph can be
large and hence loops can be present.

\begin{figure}[!t] 
\begin{center}
\includegraphics[scale=.51]{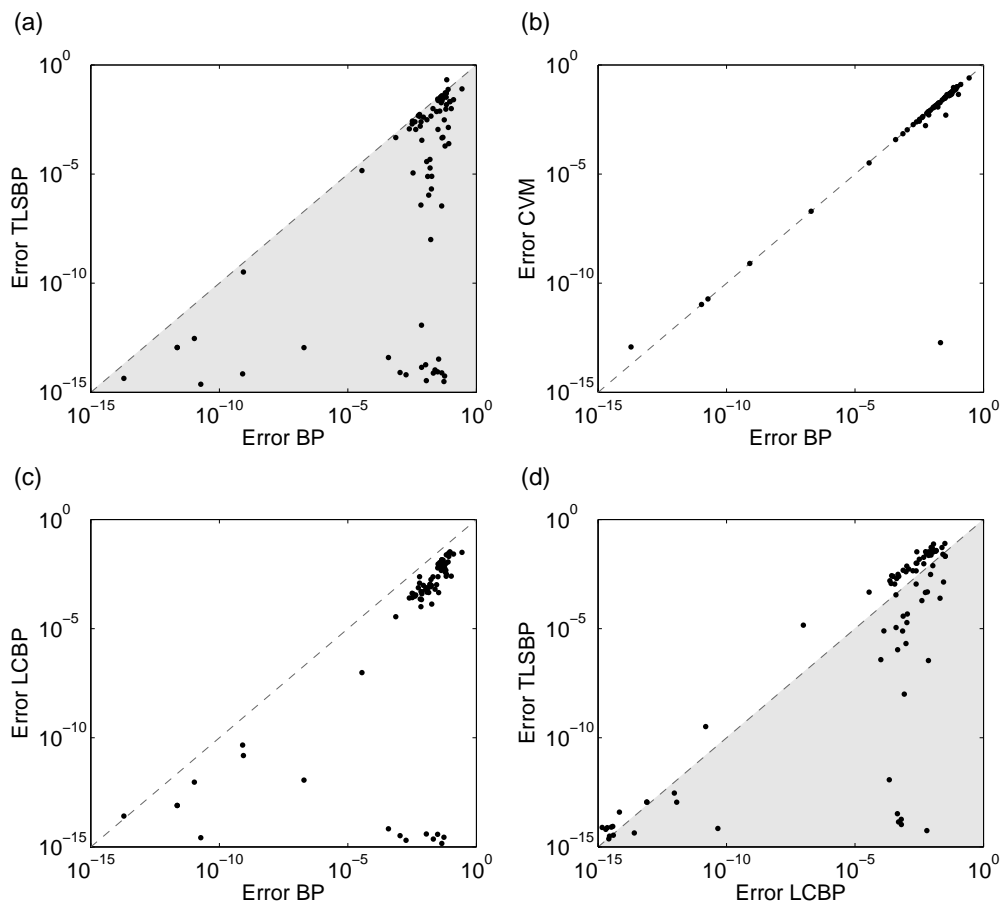} 
\end{center}
\caption{
Results of 146 random patient cases with one disease.
\textbf{(a)} TLSBP error versus BP error. \textbf{(b)}
CVM error versus BP error. \textbf{(c)} LCBP error versus BP error.
and \textbf{(d)} LCBP error versus TLSBP error.
}
\label{fig:promedas_error}
\end{figure}

In this subsection, as well as comparing TLSBP with CVM, we also use
another loop correction approach, loop corrected belief propagation
(LCBP)~\citep{MooijKappen06}, which is based on the cavity method and
also improves over BP estimates.
We use the following parameters for TLSBP: $S=100$, $M=5$, and no bound $b$.
Again, we apply the junction tree method to obtain exact marginals and
compare the different errors.
Figure \ref{fig:promedas_error} shows results for 146 different 
random instances.

We first analyze the TLSBP results compared with BP (Figure \ref{fig:promedas_error}a).
The region in light gray color indicates TLSBP improvement over BP.
The observed results vary strongly because of the wide diversity of the
particular instances, but we can basically differentiate two
scenarios.
The first set of results include those instances where the BP error is
corrected almost up to machine precision. These patient cases
correspond to graphs where exhaustive enumeration is tractable, and
TLSBP found almost all the generalized loops.
These are the dots appearing in the bottom part of Figure
\ref{fig:promedas_error}a, approximately $14\%$ of the patient
cases.
Note that even for errors of the order of $10^{-2}$ the error was
completely corrected.
Apart from these results, we observe another group of instances where
the BP error was not completely corrected.
These cases correspond to the upper dots of Figure
\ref{fig:promedas_error}a.
The results in these patient cases vary from no significant
improvements to improvements of four orders of magnitude.

Figure \ref{fig:promedas_error}b shows the 
performance of CVM considering all maximal factors together with all
loops that consist up to four different variables as outer regions.
We can see that, contrary to TLSBP, CVM in this domain performs poorly.
For only one instance
the CVM result is significantly better than BP.
Moreover, the computation time required by CVM was much larger than TLSBP in
all instances (data not shown).
These results can be complemented with the study developed in~\citep{MooijKappen06},
where it is shown that CVM does not perform significantly better
for other choices of regions.

\begin{figure}[!t] 
\begin{center}
\includegraphics[scale=.52]{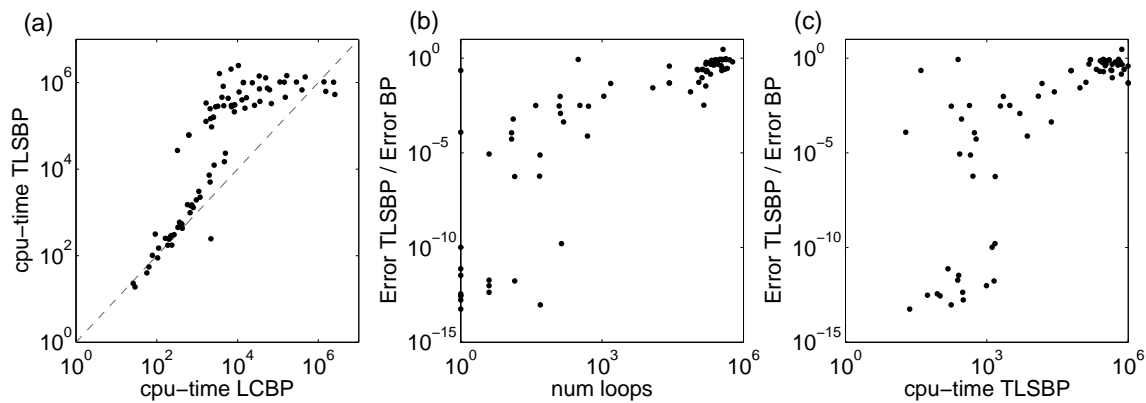} 
\end{center}
\caption{
  Results of applying TLSBP to 146 patient cases with one disease.
  \textbf{(a)} Relation between computational cost needed by LCBP and TLSBP.
  \textbf{(b)} Ratio between TLSBP and BP errors versus number of loops found.
  \textbf{(c)} Ratio between TLSBP and BP errors versus computation time.
}
\label{fig:promedas_cpu}
\end{figure}

Figure \ref{fig:promedas_error}c shows results of LCBP, the approach
presented in~\citep{MooijKappen06} on the same set of instances.
As in the case of TLSBP, LCBP significantly improves over BP.
A comparison between both approaches is illustrated in Figure \ref{fig:promedas_error}d,
where those instances where TLSBP is better are marked in light gray color.
For $41\%$ of the cases TLSBP improves the LCBP results, sometimes notably.
TLSBP enhacements were made at the cost of more time,
as Figure \ref{fig:promedas_cpu}a suggest, where in $85\%$ of the instances
TLSBP needs more time.

To analyze the TLSBP results in more depth we plot the ratio between the
error obtained by TLSBP and the BP error versus the number of
generalized loops found and the CPU time.
From Figure \ref{fig:promedas_cpu}b we can deduce that
cases where the BP error was most improved, often correspond to graphs with a
small number of generalized loops found, whereas instances with
highest number of loops considered have minor improvements.
This is explained by the fact that some instances which contained a few
loops were easy to solve and thus the BP error was significantly
reduced.
An example of one of those instances corresponds to the
Figure~\ref{fig:typicalgraphs} (left).
On the contrary, there exist very loopy instances where computing
some terms was not useful, even if a large number of them (more than
one million) where considered. A typical instance of this
type is shown in Figure~\ref{fig:typicalgraphs} (right).
The same argument is suggested by Figure \ref{fig:promedas_cpu}c
where CPU time is shown, which is often proportional to the number of loops
found.

In general, we can conclude that although the BP error was corrected in
most of the instances, there were some cases in which TLSBP did not give
significant improvements.
Considering all patient cases, the BP error was corrected in more than
one order of magnitude for more than $30\%$ of the cases.

\section{Discussion}\label{sec:discussion}
We have presented TLSBP, an algorithm to compute corrections to the BP
solution based on the loop series expansion proposed by~\citet{chertkov2006a}.
\footnote{The source code of the algorithm and a subset of the
datasets used in the experimental section can be downloaded
from : \texttt{http://www.cns.upf.edu/vicent/}.}
In general, for cases where all loops can be enumerated the method
computes the exact solution efficiently.
In contrast, if exhaustive enumeration is not tractable, the BP error
can be reduced significantly.
The performance of the algorithm does not depend directly on the size
of the problem, but on how loopy the original graph is, although for
larger instances it is more likely that more loops are present.

We have also shown that the performance of TLSBP strongly depends on
the degree of coupling between the variables.
For weak couplings, errors in BP are most prominently caused by small
simple loops and truncating the series is very useful, whereas
for strongly coupled variables, loop terms of different sizes
tend to be of the same order of magnitude, and truncating the series
can be useless.
For those difficult cases, BP convergence is also difficult, and
magnetizations at the fixed point tend to be close to extremum values,
causing numerical difficulties in the calculation of the loop expansion
formulas.
In general, we can conclude that the proposed approach is useful
in an intermediate regime, where BP results are not very accurate,
but BP is still converging.

We have confirmed empirically that there is a correlation between the BP
result and the potential improvements using TLSBP.
Those cases where the BP estimate is most corrected correspond often
to cases where the BP estimate is already accurate.
Whether a given BP error is acceptable or not depends on the inference task
and the specific domain.

The proposed approach has been compared with CVM selecting
loops of four variables and maximal factors as outer clusters.
For highly regular domains with translation invariances such as
square grids, CVM performs better than TLSBP in difficult instances
(strong interactions).
This is not surprising, since CVM exploits the symmetries on the
original graph.
However, for other domains such as random graphs,
or medical diagnosis, TLSBP show comparable, or even better results
than CVM with our choice of clusters.

The TLSBP algorithm searches the graph without
considering information accessible from the BP solution, which is used
to compute the loop corrections only as a final step.
Therefore, it can be regarded as a blind search procedure.
We have also experimented with a more ``heuristic'' algorithm where
the search is guided in some principled way.
Two modifications of the algorithm have been done in that direction:
\begin{enumerate}
\item One approach consisted in modifying the third step in a way that,
instead of applying blind mergings, generalized loops which have
larger contributions (largest $|r(C)|$) were merged preferentially.
In practice, this approach tended to check all combinations of small loops
which produced the same generalized loop, causing many redundant mergings.
Moreover, the cost of maintaining sorted the ``best'' generalized loops
caused a significant increase in the computational complexity.
This approach did not produce more accurate results neither was a more
efficient approach.
\item Also, instead of pruning the DFS search for complex loops using the
parameter $M$, we have used the following strategy:
we computed iteratively the \emph{partial term} of the loop that is being
searched, such that at each DFS step one new term using
Equations~\eqref{eq:term-var} and \eqref{eq:term-fact}
is multiplied with the current \emph{partial term}.
If at some point, the \emph{partial term} was smaller than a certain
threshold $\lambda$, the DFS was pruned. This new parameter $\lambda$
was then used instead of $M$ and result in an appropriate strategy for
graphs with weak interactions.
For cases where many terms of the same order existed, a small change
of $\lambda$ caused very different execution times, and often too deep
searches.
We concluded that using parameter $M$ is a more suitable choice in
general.
\end{enumerate}
TLSBP can be easily extended in other ways.
For instance, as an anytime algorithm.
In this context, the partition sum or marginals can be computed incrementally as
more generalized loops are being produced.
This allows to stop the algorithm at any step and presumably, the more
time used, the better the solution.
The ``improvement if allowed more time'' can be a desirable property
for applications in approximate reasoning, \citep{zilberstein95approximate}.
Another way to extend the approach is to consider the search for loops
as a \emph{compilation} stage of the inference task, which can be 
done offline.
Once all loops are retrieved and stored, the inference task would
require much less computational cost to be performed.

During the development of this work another way of selecting
generalized loops has been proposed \citep{chertkov2006b} in the
context of Low Density Parity Check codes.
Their approach tries to find only a few \emph{critical} simple loops,
related with dangerous noise configurations that lead to Linear Programing
decoding failure, and use them to modify the standard BP equations.
Their method shows promising results for the LDPC domain,
and can be applied to any general graphical model as well,
so it would be interesting to compare both approaches.

There exists another type of loop correction methods that improves BP,
which is quite different from the approach discussed here
\citep{montanari05,parisi06,MooijWemmenhoveKappenRizzo07,MooijKappen06}.
Their argument is based on the cavity method.
BP assumes that in the absence of variable $i$, the distribution of
its Markov blanket factorizes over the individual variables.
In fact, this assumption is only approximately true, due to the loops
in the graph.
The first loop correction is obtained by considering the network
with variable $i$ removed and estimating the correlations in the Markov
blanket. This argument can be applied recursively, yielding the higher
order loop corrections.
Whereas TLSBP computes exactly the corrections of a limited number of
loops, the cavity based approach computes approximately the
corrections due to all loops. An in-depth comparison of the efficiency
and accuracy of these approaches should be made.

As a final remark, we mention the relation of the loop series
expansion with a similar method originated in statistical physics, namely,
the high-temperature expansion for Ising models.
This expansion of the partition function is similar to the one
proposed by~\citet{chertkov2006a}, in the sense that every term has also
a direct diagrammatic representation on the graph, although not in terms
of generalized loops.
Note however, that the loop expansion is relative to the BP result,
contrary to the high-temperature expansion.
Another difference is
that the high temperature expansion is an expansion in a small
parameter (the inverse temperature), whereas the loop expansion has no
such small parameter.
Finally, another related approach is the walk-sum framework for inference
in certain Gaussian Markov Models~\citep{mjw_walksum_jmlr06},
where means and covariances between any two nodes of the graph
have an interpretation in terms of an expansion of walks
in the graph.
They also show that Gaussian loopy BP
has a walk-sum interpretation, computing all walks for the means but only a subset
of walks for the variances. 



\acks{
We would like to thank the reviewers for their constructive suggestions that helped us to improve the paper.
We also acknowledge financial support from the C\'atedra Telef\'onica Multim\`edia,
the Interactive Collaborative Information Systems (ICIS) project (supported by the Dutch 
Ministry of Economic Affairs, grant BSIK03024), and the Dutch Technology Foundation (STW).
Finally, we thank Bastian Wemmenhove for providing patient cases of the Promedas medical system,
and Andreas Kaltennbruner for useful suggestions.
}









\vskip 0.2in
\bibliography{jmlr-gomez}

\begin{thebibliography}{36}
\providecommand{\natexlab}[1]{#1}
\providecommand{\url}[1]{\texttt{#1}}
\expandafter\ifx\csname urlstyle\endcsname\relax
  \providecommand{\doi}[1]{doi: #1}\else
  \providecommand{\doi}{doi: \begingroup \urlstyle{rm}\Url}\fi

\bibitem[Chertkov and Chernyak()]{chertkov2006a}
M.~Chertkov and V.~Y. Chernyak.
\newblock Loop series for discrete statistical models on graphs.
\newblock \emph{Journal of Statistical Mechanics: Theory and Experiment},
  2006\penalty0 (06):\penalty0 P06009.
\newblock URL \url{http://arxiv.org/abs/cond-mat/0601487}.

\bibitem[Chertkov and Chernyak(2006)]{chertkov2006b}
M.~Chertkov and V.~Y. Chernyak.
\newblock Loop calculus helps to improve belief propagation and linear
  programming decodings of {LDPC} codes.
\newblock In \emph{invited talk at 44th Allerton Conference}, September 2006.
\newblock URL \url{http://www.arxiv.org/abs/cs/0609154}.

\bibitem[Elidan et~al.(2006)Elidan, McGraw, and Koller]{Elidanal}
G.~Elidan, I.~McGraw, and D.~Koller.
\newblock Residual belief propagation: Informed scheduling for asynchronous
  message passing.
\newblock In \emph{Proceedings of the Twenty-second Conference on Uncertainty
  in AI (UAI)}, Boston, Massachussetts, July 2006.

\bibitem[Freeman et~al.(2000)Freeman, Pasztor, and Carmichael]{freeman00}
W.~T. Freeman, E.~C. Pasztor, and O.~T. Carmichael.
\newblock Learning low-level vision.
\newblock \emph{Int. J. Comp. Vision}, 40:\penalty0 25--47, 2000.

\bibitem[Gallager(1963)]{gallager63}
R.~G. Gallager.
\newblock \emph{Low-density parity check codes}.
\newblock MIT Press, 1963.

\bibitem[Heskes et~al.(2003)Heskes, Albers, and Kappen]{hes03a}
T.~Heskes, K.~Albers, and H.~J. Kappen.
\newblock Approximate inference and constraint optimisation.
\newblock In \emph{Proceedings UAI}, pages 313--320, 2003.

\bibitem[Jaakkola and Jordan(1999)]{jaakkola_jordan99}
T.~Jaakkola and M.~I. Jordan.
\newblock Variational probabilistic inference and the {QMR-DT} network.
\newblock \emph{Journal of artificial intelligence research}, 10:\penalty0
  291--322, 1999.

\bibitem[Johnson(1975)]{Johnson75}
D.~B. Johnson.
\newblock Finding all the elementary circuits of a directed graph.
\newblock \emph{SIAM J. Comput.}, 4\penalty0 (1):\penalty0 77--84, 1975.

\bibitem[Jordan et~al.(1999)Jordan, Ghahramani, Jaakkola, and Saul]{Jordan99}
M.~Jordan, Z.~Ghahramani, T.~S. Jaakkola, and L.~Saul.
\newblock An introduction to variational methods for graphical models.
\newblock In \emph{Learning in Graphical Models}, pages 105--161. Cambridge,
  MA: MIT Press, 1999.

\bibitem[Kschischang et~al.(2001)Kschischang, Frey, and
  Loeliger]{kschischang01factor}
F.~R. Kschischang, B.~J. Frey, and H.-A. Loeliger.
\newblock Factor graphs and the sum-product algorithm.
\newblock \emph{IEEETIT: IEEE Transactions on Information Theory}, 47, 2001.

\bibitem[Lauritzen and Spiegelhalter(1988)]{lauritzen88}
S.~L. Lauritzen and D.~J. Spiegelhalter.
\newblock Local computations with probabilities on graphical structures and
  their application to expert systems.
\newblock \emph{J. Royal Statistical society B}, 50:\penalty0 154--227, 1988.

\bibitem[Leisink and Kappen(2001)]{Leisink2001}
M.~A.~R. Leisink and H.~J. Kappen.
\newblock A tighter bound for graphical models.
\newblock \emph{Neural Comput.}, 13\penalty0 (9):\penalty0 2149--2171, 2001.

\bibitem[Malioutov et~al.(2006)Malioutov, Johnson, and
  Willsky]{mjw_walksum_jmlr06}
D.~Malioutov, J.~Johnson, and A.~Willsky.
\newblock Walk-sums and belief propagation in gaussian graphical models.
\newblock \emph{Journal of Machine Learning Research}, 7, October 2006.

\bibitem[Mc{E}liece et~al.(1998)Mc{E}liece, MacKay, and Cheng]{mcelice98}
R.~Mc{E}liece, D.~MacKay, and J.~Cheng.
\newblock Turbo decoding as an instance of {Pearl's} belief propagation
  algorithm.
\newblock \emph{Journal of Selected Areas of Communication}, 16:\penalty0
  140--152, 1998.

\bibitem[M\'ezard et~al.(2002)M\'ezard, Parisi, and Zecchina]{mezard02}
M.~M\'ezard, G.~Parisi, and R.~Zecchina.
\newblock Analytic and algorithmic solution of random satisfiability problems.
\newblock \emph{Science}, 297:\penalty0 812--815, 2002.

\bibitem[Montanari and Rizzo(2005)]{montanari05}
A.~Montanari and T.~Rizzo.
\newblock How to compute loop corrections to the {Bethe} approximation.
\newblock \emph{Journal of Statistical Mechanics: Theory and Experiment},
  2005\penalty0 (10):\penalty0 P10011, 2005.
\newblock URL \url{http://arxiv.org/abs/cond-mat/0506769}.

\bibitem[Mooij and Kappen(2007)]{MooijKappen06}
J.~M. Mooij and H.~J. Kappen.
\newblock Loop corrections for approximate inference.
\newblock \emph{Journal of Machine Learning Research}, 8:\penalty0 1113--1143,
  May 2007.
\newblock URL \url{http://arxiv.org/abs/cs/0612030}.

\bibitem[Mooij and Kappen(2005)]{mooij2005}
J.~M. Mooij and H.~J. Kappen.
\newblock On the properties of the {Bethe} approximation and loopy belief
  propagation on binary networks.
\newblock \emph{Journal of Statistical Mechanics: Theory and Experiment},
  2005\penalty0 (11):\penalty0 P11012, 2005.

\bibitem[Mooij et~al.(2007)Mooij, Wemmenhove, Kappen, and
  Rizzo]{MooijWemmenhoveKappenRizzo07}
J.~M. Mooij, B.~Wemmenhove, H.~J. Kappen, and T.~Rizzo.
\newblock Loop corrected belief propagation.
\newblock In \emph{Proceedings of the Eleventh International Conference on
  Artificial Intelligence and Statistics, AISTATS}, 2007.

\bibitem[Murphy et~al.(1999)Murphy, Weiss, and Jordan]{murphy99}
K.~P. Murphy, Y.~Weiss, and M.~I. Jordan.
\newblock Loopy belief propagation for approximate inference: An empirical
  study.
\newblock In \emph{Proceedings of Uncertainty in AI}, pages 467--475, 1999.

\bibitem[Parisi and Slanina(2006)]{parisi06}
J.~Parisi and F.~Slanina.
\newblock Loop expansion around the {Bethe-Peierls} approximation for lattice
  models.
\newblock \emph{Journal of Statistical Mechanics}, page L02003, 2006.

\bibitem[Pearl(1988)]{pearl88}
J.~Pearl.
\newblock \emph{Probabilistic Reasoning in Intelligent Systems: Networks of
  Plausible Inference}.
\newblock Morgan Kaufmann Publishers Inc., San Francisco, CA, USA, 1988.
\newblock ISBN 1558604790.

\bibitem[Pelizzola(2005)]{Pelizola05}
A.~Pelizzola.
\newblock Cluster variation method in statistical physics and probabilistic
  graphical models.
\newblock \emph{Journal of Physics A: Mathematical and General}, 38\penalty0
  (33):\penalty0 R309--R339, 2005.
\newblock URL \url{http://arxiv.org/abs/cond-mat/0508216}.

\bibitem[Potamianos and Goutsias(1997)]{Potamianos1997}
G.~Potamianos and J.~Goutsias.
\newblock Stochastic approximation algorithms for partition function estimation
  of {Gibbs} random fields.
\newblock \emph{Information Theory, IEEE Transactions on,}, 43, 6:\penalty0
  1948--1965, Nov 1997.

\bibitem[Shwe et~al.(1991)Shwe, Middleton, Heckerman, Henrion, Horvitz, Lehman,
  and Cooper]{shwe91}
M.~A. Shwe, B.~Middleton, D.~E. Heckerman, M.~Henrion, E.~J. Horvitz, H.~P.
  Lehman, and G.~F. Cooper.
\newblock {Probabilistic diagnosis using a reformulation of the internist-1/
  {QMR} knowledge base}.
\newblock \emph{Methods of Information in Medicine}, 30:\penalty0 241--55,
  1991.

\bibitem[Sun et~al.(2005)Sun, Li, Kang, and Shum]{sun05}
J.~Sun, Y.~Li, S.~B. Kang, and H.~Y. Shum.
\newblock Symmetric stereo matching for occlusion handling.
\newblock \emph{Proceedings CVPR}, 2:\penalty0 399--406, 2005.

\bibitem[Takinawa and {D'Ambrosio}(1999)]{ambrosio99multiplicative}
M.~Takinawa and B.~{D'Ambrosio}.
\newblock Multiplicative factorization of noisy-{MAX}.
\newblock In \emph{Proceedings of the 15th Conference on Uncertainty in
  Artificial Intelligence}, pages 622--30, 1999.

\bibitem[Tarjan(1973)]{Tarjan73}
R.~E. Tarjan.
\newblock Enumeration of the elementary circuits of a directed graph.
\newblock \emph{SIAM J. Comput.}, 2\penalty0 (3):\penalty0 211--216, 1973.

\bibitem[Tiernan(1970)]{Tiernan70}
J.~C. Tiernan.
\newblock An efficient search algorithm to find the elementary circuits of a
  graph.
\newblock \emph{Commun. ACM}, 13\penalty0 (12):\penalty0 722--726, 1970.

\bibitem[Wainwright et~al.(2005)Wainwright, Jaakkola, and
  Willsky]{wainwright05}
M.~Wainwright, T.~Jaakkola, and A.~Willsky.
\newblock A new class of upper bounds on the log partition function.
\newblock 51 (7):\penalty0 2313--2335, July 2005.

\bibitem[Welling et~al.(2005)Welling, Minka, and Teh]{Welling05}
M.~Welling, T.~Minka, and Y.~W. Teh.
\newblock Structured region graphs: Morphing {EP} into {GBP}.
\newblock In \emph{Proceedings of the 21th Annual Conference on Uncertainty in
  Artificial Intelligence (UAI-05)}, page 609, Arlington, Virginia, 2005. AUAI
  Press.

\bibitem[Wiegerinck et~al.(1999)Wiegerinck, Kappen, ter Braak, Burg, Nijman, O,
  and Neijt]{wie99b}
W.~Wiegerinck, H.~J. Kappen, E.~W. M.~T. ter Braak, W.~J. P.~P. Burg, M.~J.
  Nijman, Y.~L. O, and J.~P. Neijt.
\newblock Approximate inference for medical diagnosis.
\newblock \emph{Pattern Recognition Letters}, 20:\penalty0 1231--1239, 1999.

\bibitem[Yedidia et~al.(2001)Yedidia, Freeman, and Weiss]{yedidia00}
J.~S. Yedidia, W.~T. Freeman, and Y.~Weiss.
\newblock Generalized belief propagation.
\newblock In T.K. Leen, T.G. Dietterich, and V.~Tresp, editors, \emph{Advances
  in Neural Information Processing Systems 13 (Proceedings of the 2000
  Conference)}, 2001.

\bibitem[Yedidia et~al.(2005)Yedidia, Freeman, Weiss, and Yuille]{yedidia05}
J.~S. Yedidia, W.~T. Freeman, Y.~Weiss, and A.~L. Yuille.
\newblock Constructing free-energy approximations and generalized belief
  propagation algorithms.
\newblock \emph{Information Theory, IEEE Transactions on}, 51:\penalty0
  2282--2312, 2005.

\bibitem[Yuille(2002)]{yuille02}
A.~L. Yuille.
\newblock {CCCP} algorithms to minimize the {Bethe} and {Kikuchi} free
  energies: Convergent alternatives to belief propagation.
\newblock \emph{Neural computation}, 14:\penalty0 1691--1722, 2002.

\bibitem[Zilberstein and Russell(1996)]{zilberstein95approximate}
S.~Zilberstein and S.~Russell.
\newblock Approximate reasoning using anytime algorithms.
\newblock \emph{AI magazine}, 17\penalty0 (3):\penalty0 73--83 (1p.1/4), 1996.
\newblock ISSN 0738-4602.

\end{thebibliography}

\newpage

\end{document}